\PassOptionsToPackage{table}{xcolor}
\documentclass[fleqn,10pt]{wlscirep}
\usepackage[utf8]{inputenc}
\usepackage[T1]{fontenc}
\usepackage{helvet}
\usepackage{caption}
\usepackage{subcaption}
\title{Generative Capacity of Probabilistic Protein Sequence Models}

\author[1,2,4]{Francisco McGee}
\author[2,5]{Quentin Novinger}
\author[1,3,4,6]{Ronald M Levy}
\author[2,3,*]{Vincenzo Carnevale}
\author[1,6,*]{Allan Haldane}
\affil[1]{Center for Biophysics and Computational Biology, Temple University, Philadelphia, 19122, USA}
\affil[2]{Institute for Computational Molecular Science, Temple University, Philadelphia, 19122, USA}
\affil[3]{Department of Biology, Temple University, Philadelphia, 19122, USA}
\affil[4]{Department of Chemistry, Temple University, Philadelphia, 19122, USA}
\affil[5]{Department of Computer \& Information Sciences, Temple University, Philadelphia, 19122, USA}
\affil[6]{Department of Physics, Temple University, Philadelphia, 19122, USA}
\affil[*]{Corresponding authors: vincenzo.carnevale@temple.edu, allan.haldane@temple.edu}

\newcommand{\red}[1]{{#1}}
\newcommand{\redb}[1]{{#1}}

\begin{document}

\begin{abstract}
\red{

Potts models and variational autoencoders (VAEs) have recently gained popularity as generative protein sequence models (GPSMs) to explore fitness landscapes and predict the effect of mutations.  Despite encouraging results, quantitative characterization and comparison of GPSM-generated probability distributions is still lacking. It is currently unclear whether GPSMs can faithfully reproduce the complex multi-residue mutation patterns observed in natural sequences arising due to epistasis. \redb{We develop a set of sequence statistics to assess the “generative capacity” of three GPSMs of recent interest: the pairwise Potts Hamiltonian, the VAE, and the site-independent model, using natural and synthetic datasets.}  We show that the generative capacity of the Potts Hamiltonian model is the largest; the higher order mutational statistics generated by the model agree with those observed for natural sequences. In contrast, we show that the VAE's generative capacity lies between the pairwise Potts and site-independent models. \redb{Importantly, our work measures GPSM generative capacity in terms of higher-order sequence covariation statistics which we have developed, and provides a new framework for evaluating and interpreting GPSM accuracy that emphasizes the role of epistasis.}}

\end{abstract}

\flushbottom
\maketitle
\section*{Introduction}
Recent progress in decoding the patterns of mutations in protein multiple sequence alignments (MSAs) has highlighted the importance of mutational covariation in determining protein function, conformation and evolution, and has found practical applications in protein design, drug design, drug resistance prediction, and classification~\cite{Levy2017,cocco_inverse_2018,tubiana_learning_2019}. These developments were sparked by the recognition that the pairwise covariation of mutations observed in large MSAs of evolutionarily diverged sequences belonging to a common protein family can be used to fit maximum entropy ``Potts'' statistical models~\cite{lapedes_correlated_1999,weigt_identification_2009,haldane_structural_2016}. These contain pairwise statistical interaction parameters reflecting epistasis~\cite{domingo_causes_2019} between pairs of positions. Such models have been shown to accurately predict physical contacts in protein structure~\cite{haldane_structural_2016,noel_sequence_2016,morcos_coevolutionary_2013,sulkowska_genomics-aided_2012}, and have been used to significantly improve the prediction of the fitness effect of mutations to a sequence compared to site-independent sequence variation models which do not account for covariation~\cite{hopf_mutation_2017,biswas_epistasis_2019}. They are ``generative'' in the sense that they define the probability, $p(S)$, that a protein sequence $S$ results from the evolutionary process. Intriguingly, the probability distribution $p(S)$ can be used to sample unobserved, and yet viable, artificial sequences. In practice, the model distribution $p(S)$ depends on parameters that are found by maximizing a suitably defined likelihood function on observations provided by the MSA of a target protein family. As long as the model is well specified and generalizes from the training MSA, it can then be used to generate new sequences, and thus a new MSA whose statistics should match those of the original target protein family. We refer to probabilistic models that create new protein sequences in this way as generative protein sequence models (GPSMs).

The fact that Potts maximum entropy models are limited to pairwise epistatic interaction terms and have a simple functional form for $p(S)$ raises the possibility that their functional form is not flexible enough to describe the data, i.e. that the model is not well specified. While a model with only pairwise interaction terms can predict complex patterns of covariation involving three or more positions through chains of pairwise interactions, it cannot model certain triplet and higher patterns of covariation that require a model with more than pairwise interaction terms~\cite{Bialek2003}. For example, a Potts model cannot predict patterns described by an XOR or boolean parity function in which the $n$-th residue is determined by whether an odd number of the $n-1$ previous residues have a certain value (see Supplementary Information). While some evidence has suggested that in the case of protein sequence data the pairwise model is sufficient and necessary to model sequence variation~\cite{bialek_rediscovering_2007,Weigt2018,haldane_coevolutionary_2018}, some of this evidence is based on averaged properties, and there appears to be some weak evidence for the possibility of rare ``higher-order epistasis'' affecting protein evolution~\cite{weinreich_should_2013,domingo_causes_2019,Haq2012,Haq2009}, by which we mean the possibility that subsequence frequencies of three or more positions cannot be reproduced by a model with only pairwise interactions. Fitting maximum entropy models with all triplet interactions is not feasible without significant algorithmic innovation, since for a protein of length 100 it would require approximately 10B parameters and enormous MSA datasets to overcome finite sampling error (see Supplementary Information). However, recent developments in powerful machine learning techniques applied to images, language, and other data have shown how complex distributions $p(S)$ can be fit with models using more manageable parameter set sizes. Building on the demonstrated power of incorporating pairwise epistasis into protein sequence models, this has motivated investigation of machine learning strategies for generative modelling of protein sequence variation which can go beyond pairwise interactions, including Restricted Boltzmann Machines (RBMs)~\cite{tubiana_learning_2019}, variational autoencoders (VAEs)~\cite{riesselman_deep_2018,ding_deciphering_2019,sinai_variational_2018,costello_how_2019}, General Adversarial Networks (GANs)~\cite{gupta_feedback_2019}, transformers~\cite{madani_progen_2020,vig2021bertology,elnaggar_prottrans_2020,choromanski_rethinking_2021}, and others. 

One model in particular, the VAE~\cite{kingma_auto-encoding_2014,rezende_stochastic_2014}, has been cited as being well suited for modelling protein sequence covariation, with the potential to detect higher order epistasis~\cite{riesselman_deep_2018,ding_deciphering_2019}. The VAE also potentially gives additional insights into the topology of protein sequence space through examination of the ``latent'' (hidden) parameters of the model, which have been suggested to be related to protein sequence phylogenetic relationships~\cite{riesselman_deep_2018,ding_deciphering_2019,sinai_variational_2018,elnaggar_prottrans_2020}. One implementation of a VAE-GPSM, ``DeepSequence'', found that the VAE model was better able to predict experimental measurements of the effect of mutations \red{reported in deep mutational scans} than a pairwise Potts model, which was attributed to the VAE's ability to model higher-order epistasis~\cite{riesselman_deep_2018,hopf_mutation_2017}. However, it has also been suggested by others that the improvement reported for DeepSequence could be attributed to the use of biologically motivated priors and engineering efforts, rather than because it truly captured higher-order epistasis~\cite{ding_deciphering_2019}. Furthermore, while VAE-GPSMs are generative and aim to capture the protein sequence distribution $p(S)$, to our knowledge none of these studies have thoroughly tested what we will call the ``generative capacity''~\cite{luce_handbook_1963,frawley_international_2003} of the VAE model, meaning the ability of the model to generate new sequences drawn from the model distribution $p(S)$, which are statistically indistinguishable from those of a given ``target'' protein family. Testing the generative capacity, specifically higher-order covariation, of a GPSM is a fundamental check of whether the model is well specified and generalizes from the training set, two prerequisites to capturing higher-order epistasis.

Here, we perform a series of numerical experiments that explore the generative capacity of GPSMs. \redb{These GPSMs include a pairwise Potts Hamiltonian model with pairwise interaction terms (Mi3)~\cite{haldane_mi3-gpu_2020}, two state-of-the-art implementations of a variational autoencoder (VAE), and a site-independent model which does not model covariation (Indep). One VAE implementation is a ``standard" VAE implementation (sVAE), which uses an architecture very similar to that of EVOVAE\cite{sinai_variational_2018} recently used to model protein sequences, though optimized with a larger number of latent dimensions, and which more closely follows the VAE inference method as it was originally presented~\cite{kingma_auto-encoding_2014,rezende_stochastic_2014}. The second VAE implementation is ``DeepSequence'', mentioned above, which uses a deeper neural network and a more sophisticated optimizer (see Methods)~\cite{riesselman_deep_2018}.}

\redb{We evaluate the generative capacity of these models using three standard MSA statistics, and introduce a fourth and novel metric based on averaged higher-order marginals ($r_{20}$)~\cite{haldane_coevolutionary_2018}. The three standard metrics are pairwise covariance correlations~\cite{cocco_inverse_2018,haldane_coevolutionary_2018,shimagaki_selection_2019,hawkins-hooker_generating_2021}, Hamming distance distributions~\cite{cocco_inverse_2018,haldane_coevolutionary_2018,facco_intrinsic_2019,granata_accurate_2016}, and statistical energy correlations~\cite{haldane_coevolutionary_2018,riesselman_deep_2018,riesselman_accelerating_2019}.} We also test how each GPSM behaves as fewer training sequences are provided, and thus how well the GPSMs generalize, using two training MSA sizes: one representing the estimated size upper bound for a Pfam protein family (10K) (see Supplementary Information)~\cite{el-gebali_pfam_2019}, and one being large enough to eliminate out-of-sample error in our generative capacity measurements (see Results)~\cite{alquraishi_proteinnet_2019}. We evaluate GPSM performance both for natural datasets and for synthetic datasets in which the ground truth is known. Our comparative analysis thus consists of four intersecting variables: three GPSMs, a suite of four generative capacity metrics, two training MSA sizes, and two dataset types. 

\redb{Our results show that while VAEs model some epistasis they are unable to reproduce MSA statistics as well as Mi3, but are more accurate than Indep, which cannot model covariation, for all metrics tested.} \redb{The two VAE implementations we tested perform similarly to each other across all metrics. We find that the VAEs perform more similarly to Mi3 than they do to Indep, but perform more like the Indep model for the $r_{20}$ metric than does Mi3, which is the metric that most directly tests a GPSM's ability to model higher-order sequence covariation.} This result suggests that $r_{20}$ is a more sensitive measure of GPSM generative capacity than the other standard metrics, representing a novel and powerful metric to discriminate between GPSMs by averaging over higher-order covariation terms up to the 10$^{th}$ order. Our work consolidates several benchmark statistics of generative capacity for GPSMs and introduces a new one, $r_{20}$, offering a novel framework for evaluating and interpreting GPSM accuracy in the context of higher-order covariation. By quantifying and comparing GPSM performance in our innovative epistasis-oriented approach, we hope to better understand the challenges and limitations inherent to generative modelling of natural protein sequence datasets, better gauge the state of the art, and provide insight for future efforts in terms of minimizing the confounding effects of data limitations in generative protein sequence modeling.

\section*{Results}

\subsection*{Target Distributions}

Our goal is to set baseline expectations for the generative capacity of the GPSMs when fit to synthetic or natural protein sequence data of varying training MSA sizes. Generative models of protein MSAs define a distribution $p_\theta(S)$ for the probability of a sequence $S$ appearing in an MSA dataset given model parameters $\theta$. The model parameters are fit by either exact or approximate maximum likelihood inference of the likelihood $\mathcal{L} = \sum_{S \in \text{MSA}} p_\theta(S)$ over a training MSA, using regularization techniques to prevent overfitting. The sequences in the training MSA are assumed to be identical independent samples from a ``target'' probability distribution $p^0(S)$, which is generally unknown~\cite{kingma_auto-encoding_2014}. For a model with high generative capacity, $p_\theta(S)$ will closely approximate $p^0(S)$~\cite{ding_deciphering_2019}. We test three GPSMs: a pairwise Potts model (Mi3)~\cite{haldane_mi3-gpu_2020}; two implementations of a VAE, which are a standard VAE (sVAE) and DeepSequence~\cite{riesselman_deep_2018}; and a site-independent model (Indep), each with a different functional form of $p_\theta(S)$.

It is not possible to measure the similarity of $p_\theta(S)$ and $p^0(S)$ directly because of the high dimensionality of sequence space, since the number of sequence probabilities to compare is equal to $q^L$, where $L$ is the sequence length (typically $\sim$300) and $q$ is the alphabet size ($\sim$21). Instead, we measure how derived statistics computed from evaluation MSAs generated by each GPSM match those of target MSAs drawn from the target distribution. Three of these are standard metrics: the pairwise Hamming distance distribution, the pairwise covariance scores, and the GPSM's ability to predict $p^0(S)$ for individual sequences, also called statistical energy (see Methods). We present a fourth metric, the averaged higher order marginal accuracy $r_{20}$. 

\subsection*{GPSM Error and Experiments}

\begin{figure*}
\centering
\includegraphics[width=\linewidth]{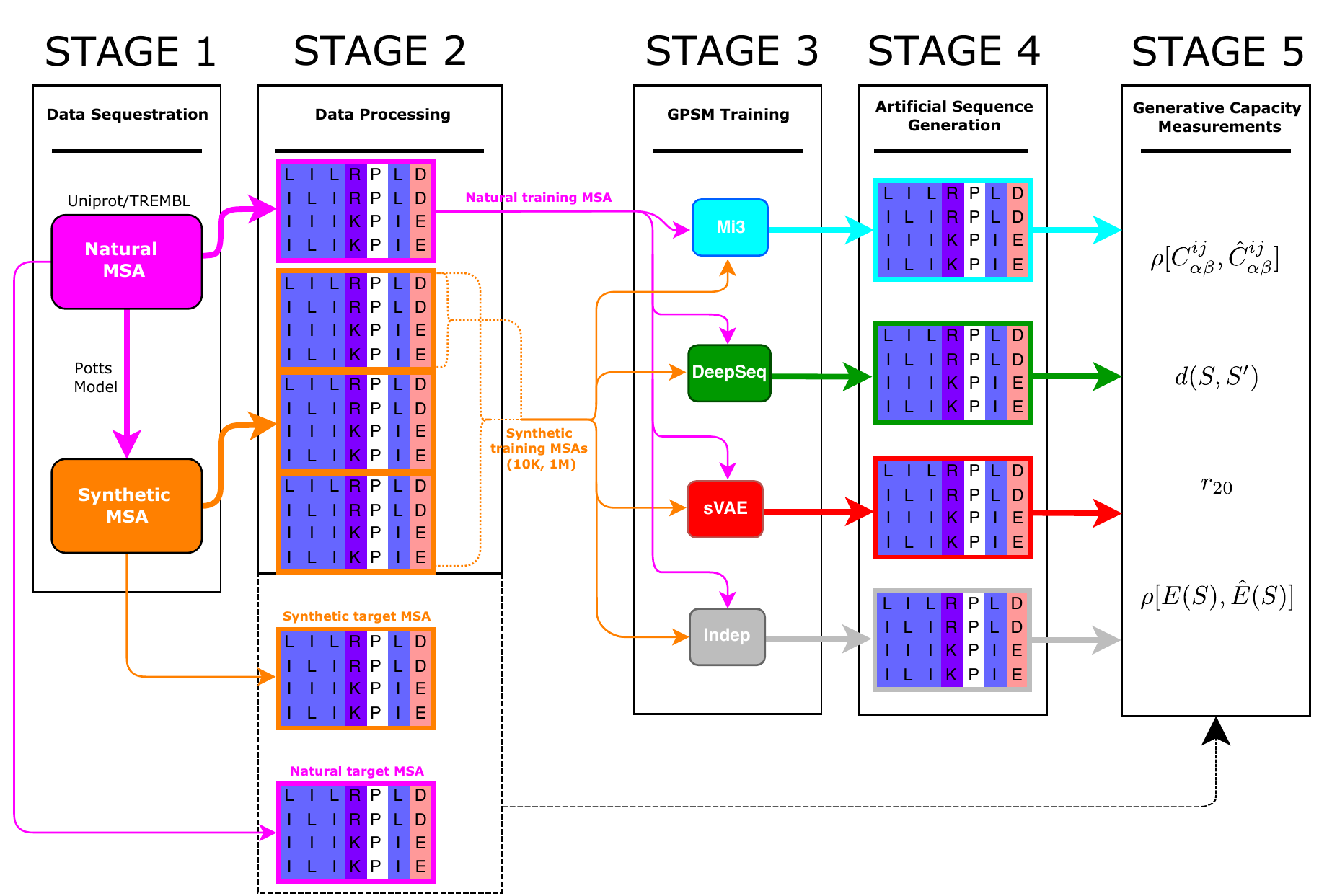}
\caption{\textbf{5-stage analysis pipeline diagram.} Stage 1: Data Sequestration. Two different MSAs are sequestered: one natural from Uniprot/TREMBL; and the other synthetic, generated by a Potts model fit to the natural MSA. Stage 2: Data Processing. Sequences are indexed and split. Non-overlapping training, target, and evaluation MSAs are shown. Stage 3: GPSM Training. GPSMs are trained. Stage 4: Artificial Sequence Generation. Evaluation MSAs are generated from each GPSM. Stage 5: Generative Capacity Measurements. Computation and visualization of generative capacity metrics is performed.}
\label{fig:pipeline}
\end{figure*}

\begin{table}
    \centering
    \rowcolors{2}{gray!25}{white}
    \begin{tabular}{|c|p{0.75\linewidth}|}
    \hline \rowcolor{gray!25}
Specification Error & 
occurs when the functional form $p_\theta(S)$ of a GPSM is not flexible enough to accurately model the target distribution $p^0(S)$ for any choice of parameters $\theta$ \\
Out-of-Sample Error &
occurs when a GPSM fit to a finite training dataset fails to correctly model unseen data, and is a consequence of overfitting \\
Estimation Error &
occurs due to statistical error in the MSA evaluation metrics when computed from finite evaluation and target MSAs \\
\hline
Training MSA & drawn from $p^0(S)$, used to train and parameterize a GPSM \\
%Target MSA & drawn from $p^0(S)$ used to evaluate the GPSM, non-overlapping with training MSA \\
Target MSA & drawn from $p^0(S)$ separately from the training MSA, used as a validation dataset for performing metrics \\
Evaluation MSA &  drawn from $p_\theta(S)$ for a parameterized GPSM,  used to compare to the target MSA \\
\hline
    \end{tabular}
    \caption{Glossary of error types, and the MSA datasets we use to evaluate these errors.}
    \label{tab:glossary}
\end{table}
  
We divide our tests into two analysis tracks: one synthetic, in which we train the GPSMs on synthetic MSAs generated from a known target distribution; and one natural, in which we train the GPSMs on a representative natural protein family MSA, the kinase super family, sequestered from Uniprot/TrEMBL~\cite{the_uniprot_consortium_uniprot_2019}. These analysis tracks are meant to probe and isolate two distinct forms of error which may cause $p_\theta(S)$ to deviate from $p^0(S)$ (see Table \ref{tab:glossary}).  The first is ``specification error''~\cite{burnham_model_2002}, which occurs when the functional form of $p_\theta(S)$ of a model is not flexible enough to accurately model the target probability distribution $p^0(S)$ for any choice of parameters. A key motivation for choosing a VAE over a Potts model is its potentially lower specification error when higher-order epistasis is present~\cite{riesselman_deep_2018}. Indeed, Potts models are limited to pairwise interaction terms of a particular functional form, while VAEs are not. The second form of error is ``out-of-sample error''~\cite{noauthor_yaser_nodate}, caused by a paucity of training samples, and is the consequence of overfitting~\cite{everitt_cambridge_2010}. Even a well-specified model could fail to generalize when fit to a small training dataset, and may mis-predict $p^0(S)$ for test sequences, so it follows that increasing training MSA size reduces out-of-sample error. Beyond specification and out-of-sample error, which each reflect an aspect of GPSM generalization error, there can be ``estimation error'' in our MSA test statistics due to the finite MSA sizes we use to estimate their values, which sets an upper bound on how well these statistics can match their target values, depending on the metric~\cite{mohri_foundations_2018}. Finally, other errors may arise due to implementation limitations of the inference methods, for instance due to finite precision arithmetic or to finite sample effects when Monte Carlo methods are used. 

The synthetic analysis allows us to isolate specification error, and minimize both out-of-sample and estimation error, because here the target distribution is known exactly, and we can generate arbitrarily large training, target, and evaluation datasets. The synthetic analysis also allows us to quantify out-of-sample error by modulating the training MSA size. We specify the synthetic target probability distribution $p^0(S)$ to be exactly the Potts model distribution we inferred based on natural protein kinase sequence data using Mi3 in our natural analysis (see Methods)~\cite{haldane_mi3-gpu_2020}. The sequences generated from this synthetic target distribution should have statistical properties similar to real, or ``natural'', protein family MSAs, albeit constrained by the fact that the Hamiltonian model used to generate the synthetic dataset is limited to pairwise epistatic interaction terms only~\cite{haldane_structural_2016, haldane_coevolutionary_2018, haldane_influence_2019}. Whereas Indep and the VAEs may still fail to model this known probability distribution in the synthetic analysis, our expectation is that Mi3 will be unaffected by specification error in the synthetic tests, since the target MSA is sampled from the same probability distribution used to carry out the inference. \red{In Supplementary Information, we also perform an alternate synthetic test which does not favor Mi3 in this way, in which the target distribution is instead specified by sVAE, finding that both Mi3 and sVAE are able to fit this target sVAE distribution accurately.} In our synthetic analysis, we test two synthetic training MSA sizes: (i) 1M sequences, to minimize overfitting effects and consequently out-of-sample error, thereby isolating GPSM specification error in this experiment; and (ii) 10K sequences, to illustrate the expected GPSM performance on typical datasets, as most protein families in Pfam have less than 10K independent effective sequences (see Supplementary Information)~\cite{el-gebali_pfam_2019}.

The natural analysis examines the performance of the models on natural sequence data, which potentially contain higher-order correlations that require a Hamiltonian model with triplet or higher-order interaction terms to capture. On this dataset, the VAEs could potentially outperform Mi3, depending on the importance of higher-order epistatic terms, if present~\cite{riesselman_deep_2018}. However, unlike in the synthetic analysis, here we do not know \textit{a priori} the target distribution and, most importantly, we have only limited datasets for both training and evaluation. In the natural tests, the training and target MSAs each contain $\sim$10K non-overlapping kinase sequences from the Uniprot/TREMBL database after phylogenetic filtering at 50\% sequence identity. \redb{After this filtering we consider each sequence to be an independent sample of the evolutionary process, which has the equilibrium distribution $p^0(S)$.}

Our overall testing procedure is outlined in Fig.\ref{fig:pipeline} (see Methods), and the terms discussed are summarized in Table \ref{tab:glossary}. Our training datasets are either a natural protein sequence dataset obtained from Uniprot/TREMBL, or a synthetic training dataset. We then fit the GPSMs to the training datasets, and generate evaluation MSAs from each model. Finally, using our suite of four generative capacity metrics, we compare statistics of the evaluation MSAs to those of ``target'' MSAs, which contain sequences drawn from the target distributions and therefore represent our expectation.

\subsection*{Pairwise covariance correlations}

\begin{figure}
    \centering
    \begin{subfigure}[b]{0.3\textwidth}
        \centering
        \caption{}
        \label{fig:1M_synth_covars}
        \includegraphics[width=\textwidth]{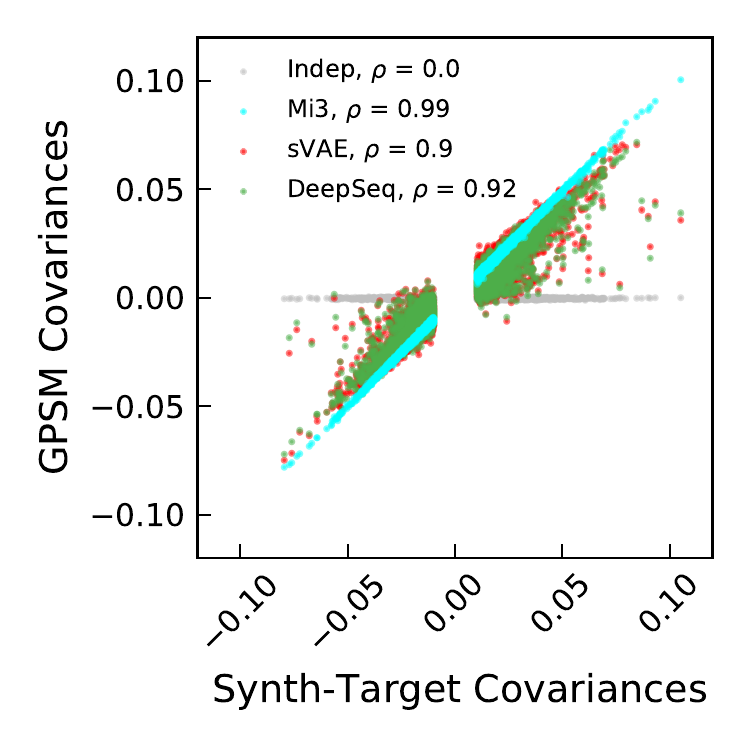}
    \end{subfigure}
    \hfill
    \begin{subfigure}[b]{0.3\textwidth}
        \centering
        \caption{}
        \label{fig:10K_synth_covars}
        \includegraphics[width=\textwidth]{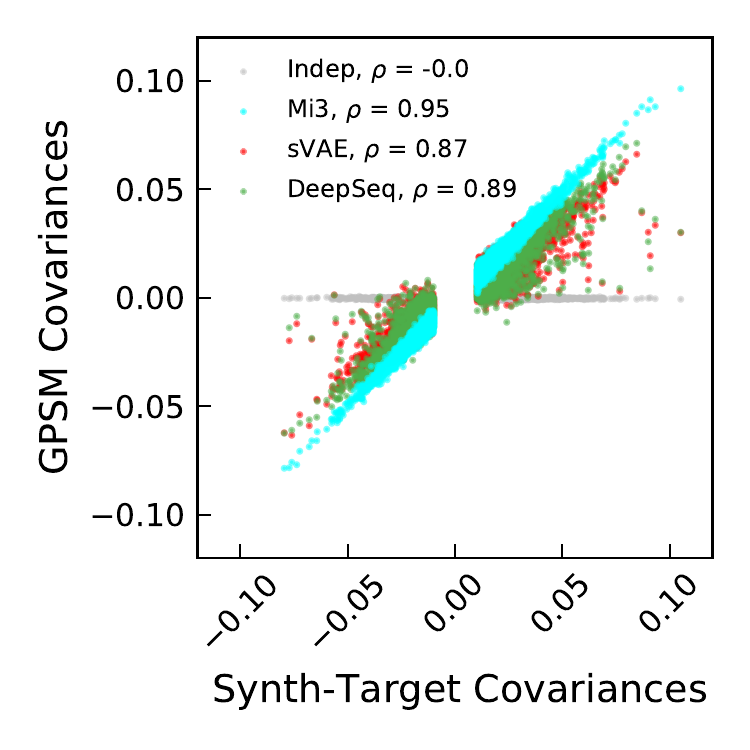}
    \end{subfigure}
    \hfill
    \begin{subfigure}[b]{0.3\textwidth}
        \centering
        \caption{}
        \label{fig:10K_nat_covars}        
        \includegraphics[width=\textwidth]{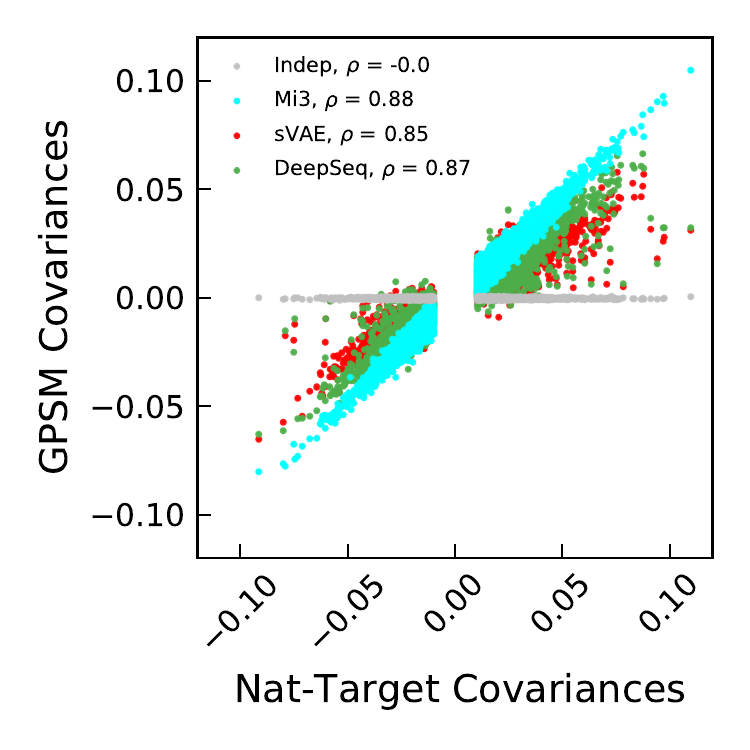}
    \end{subfigure}
    \hfill
    \begin{subfigure}[b]{0.3\textwidth}
        \centering
        \caption{}
        \label{fig:1M_synth_r20}
        \includegraphics[width=\textwidth]{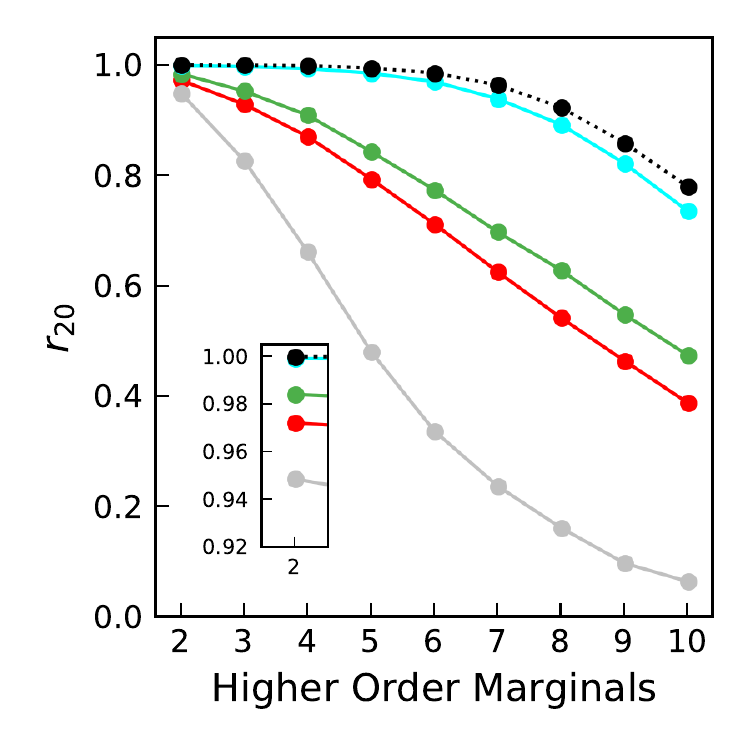}
    \end{subfigure}
    \hfill
    \begin{subfigure}[b]{0.3\textwidth}
        \centering
        \caption{}
        \label{fig:10K_synth_r20}
        \includegraphics[width=\textwidth]{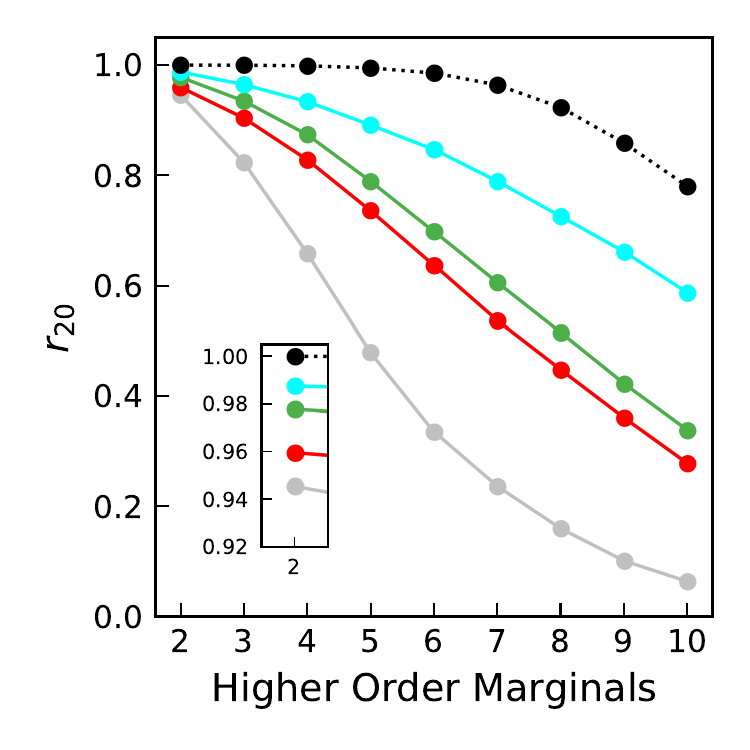}
    \end{subfigure}
    \hfill
    \begin{subfigure}[b]{0.3\textwidth}
        \centering
        \caption{}
        \label{fig:10K_nat_r20}        
        \includegraphics[width=\textwidth]{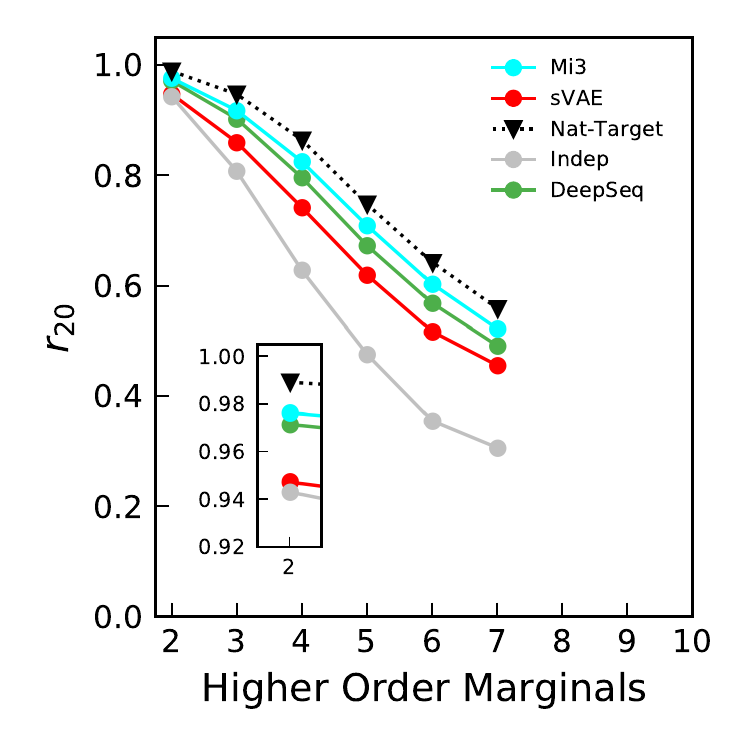}
    \end{subfigure}
    \caption{\redb{\textbf{Pairwise covariance correlations (Top Row) and Pearson $r_{20}$ correlations (Bottom Row).} GPSM MSA statistics are compared to the corresponding target values. GPSMs were trained on 1M synthetic (\textbf{a}, \textbf{d}), 10K synthetic (\textbf{b}, \textbf{e}), or 10K natural (\textbf{c}, \textbf{f}) kinase sequences. \textbf{Top Row} Covariances $C^{ij}_{\alpha\beta}$ computed from MSAs of 500K sequences generated by each GPSM (y-axis) vs target covariances (x-axis) computed from MSAs of 500K sythnetic target sequences (\textbf{a}, \textbf{b}) or from 10K natural target sequences (\textbf{c}), with Pearson correlation $\rho$ shown for each comparison. Generative capacity slightly decreases as synthetic training sample size is reduced from 1M in \textbf{a} to 10K in \textbf{b} for all GPSMs, except Indep. Covariances around zero are omitted for simplicity. \textbf{Bottom Row} Pearson $r_{20}$ score (y-axis) as a function of Higher-Order-Marginal (HOM) $n$-tuple length (x-axis), illustrating how well each GPSM predicts HOMs from the target distribution. Lines are colored as in the top panels. $r_{20}$ scores for each GPSM are computed using GPSM-generated evaluation MSAs of 6M sequences compared to target MSAs of 6M sequences for the synthetic tests (\textbf{d}, \textbf{e}) or to a target MSA of 10K natural sequences for the natural test (\textbf{f}) due to limited natural sequence data. The black dotted line denotes an estimation upper limit caused by finite sampling estimation error. This is computed as the $r_{20}$ scores for two non-overlapping synthetic target MSAs of 6M sequences each (\textbf{d} and \textbf{e}, black circles) or for MSAs of 6M and 10K sequences both generated from the Mi3 model trained on natural sequences (\textbf{f}, black triangles). For panel \textbf{f}, only lengths two through seven are plotted, as the small dataset size limits HOM estimation. Insets emphasize pairwise $r_{20}$. \red{Comparing panels \textbf{d} and \textbf{e}, the generative capacity of all GPSMs decreases with synthetic training sample size according to $r_{20}$.}}}
    \label{fig:covars_r20}        
\end{figure}

We first examine the pairwise covariance scores for pairs of amino acids of an MSA defined as $C^{ij}_{\alpha\beta} =f^{ij}_{\alpha\beta} -f^i_\alpha f^j_\beta$. Here, $f^{ij}_{\alpha\beta}$ are the MSA bivariate marginals, meaning the frequency of amino acid combination $\alpha,\beta$ at positions $i, j$ in the MSA. $f^i_\alpha$ and $f^j_\beta$ are the univariate marginals, or individual amino acid frequencies at positions $i$ and $j$. Each covariance term measures the difference between the joint frequency for pairs of amino acids and the product of the single-site residue frequencies, i.e. the expected counts in the hypothesis of statistical independence. The scores equal 0 if the two positions do not covary. Coevolving amino acids are an important aspect of sequence variation in protein MSAs, and a GPSM's ability to reproduce the pairwise covariance scores of the training dataset has been used in the past as a fundamental, non-trivial measure of the GPSM's ability to model protein sequence covariation~\cite{cocco_inverse_2018,haldane_coevolutionary_2018,shimagaki_selection_2019,hawkins-hooker_generating_2021}.

For each GPSM, we compare pairwise covariance scores for all pairs of positions and residues $\hat{C}^{ij}_{\alpha\beta}$ in their respective evaluation MSA to the corresponding target pair $C^{ij}_{\alpha\beta}$ in the target MSA using the Pearson correlation coefficient $\rho(\{C^{ij}_{\alpha\beta}\}, \{\hat{C}^{ij}_{\alpha\beta}\})$ (Fig. \ref{fig:covars_r20}, Top Row). In the synthetic tests we evaluate this statistic using 500K sequences for both the target and evaluation MSAs, while for the natural test we compare 500K evaluation sequences to the available 10K target sequences. Mi3 accurately reproduces the target covariance scores in all tests ($\rho = 0.99$ in Fig.~\ref{fig:1M_synth_covars}, $\rho = 0.95$ in Fig.~\ref{fig:10K_synth_covars}, and $\rho = 0.88$ in Fig.~\ref{fig:10K_nat_covars} respectively). The somewhat lower value for Mi3 in the natural analysis of $\rho = 0.88$ is accounted for entirely by increased estimation error in that test, as only 10K target sequences are available for evaluation, and the expected $\rho$ due only to estimation error is $\rho \sim$0.87, which is computed by comparing the \red{natural 10K target sequences to the natural 10K training sequences using this metric.} The high generative capacity of Mi3 is expected, because Mi3 parameters are optimized to exactly reproduce the joint frequencies of pairs of amino acids from the target MSA. The VAEs' inference does not include this constraint, but we find that even \red{when trained on the larger} (1M) dataset of synthetic sequences, the covariances \red{computed from the VAEs' evaluation MSAs} are generally similar to each other, and smaller in magnitude than those of the target, showing smaller correlation with the target than Mi3 ($\rho = 0.9$ for sVAE and $\rho = 0.92$ for DeepSequence in Fig.~\ref{fig:1M_synth_covars}). \red{For the VAEs, this amount of error in $\rho$ can primarily be attributed to specification error, since training GPSMs on 1M sequences largely eliminates out-of-sample error, and the large evaluation MSAs make estimation error negligible.} The VAEs' covariances are further scaled down slightly when fit to the synthetic 10K dataset ($\rho = 0.87$ for sVAE and $\rho = 0.89$ for DeepSequence in Fig.~\ref{fig:10K_synth_covars}). Indep cannot reproduce covariances by definition, so $\rho$ is zero in all tests, as expected. The generative capacity trends for this metric are consistent between the synthetic and natural analyses for all GPSMs, showing the behavior is not due to artificial properties of our synthetic target model.

These results confirm that VAEs can model pairwise epistasis in protein sequence datasets, since they generate pairwise mutational covariances that are correlated with the target values, even in the absence of explicit constraints for reproducing these statistics. However, they scale down the strength of pairwise covariances in both the synthetic and natural analyses and the correlation with the target is lower than 1. Mi3, in contrast, is constrained by design to fit the pairwise covariance scores and does so nearly perfectly.

\subsection*{Higher order marginal statistics}

A more stringent test of GPSM generative capacity is to measure the model’s ability to reproduce sequence covariation involving more than two positions, or higher-order covariation. We characterize these higher-order covariation patterns in the target MSA and GPSM-generated evaluation MSAs by computing the frequency of non-contiguous amino acid $n$-tuples, or higher-order marginals (HOMs) corresponding to subsequences, and compare their frequency in each MSA to corresponding values in the target MSAs. For increasing values of $n$ the number of possible $n$-tuple combinations increases rapidly, requiring increasingly large evaluation MSAs to accurately estimate the frequency of individual $n$-tuples. For this reason, we limit $n$-tuple length to $n \le 10$ and only compute a limited subset of all possible position sets for each $n$. \red{For each $n$ we randomly choose 3K position sets, compute the frequencies of the top twenty most frequent $n$-tuples for each corresponding position set in the target and evaluation MSAs, as these are well sampled, and for each position set compute the Pearson correlation between these top twenty frequencies.} We then average the correlation values for each $n$ over all position sets. We call this metric the Pearson correlation $r_{20}$~\cite{haldane_coevolutionary_2018}. In this test, estimation error is non-zero because of the extremely large MSAs required to compute $n$-tuple frequencies, particularly for high $n > 5$ (see Supplementary Information). We can predict the estimation error caused by finite sampling in the evaluation MSAs by computing the $r_{20}$ scores between two non-overlapping MSAs generated by the synthetic target model, which are of the same size as our evaluation MSAs.

In Fig.~\ref{fig:covars_r20}, Bottom Row, we plot the HOM $r_{20}$ for varying $n$. The expected estimation error (black line) represents a generative capacity upper bound, giving the highest measurable $r_{20}$ given the evaluation MSA size of 6M for the synthetic analysis and 10K for the natural analysis. The $r_{20}$ for Mi3 fit to 1M training sequences is very close to the validation upper-bound for all $n$, suggesting it has accurately fit the synthetic target distribution and its specification error is close to zero (Fig.~\ref{fig:1M_synth_r20}). This is expected since the synthetic target model in this test is a Potts model. With 10K training sequences, Mi3 $r_{20}$ scores are lower than the 1M result for all $n$ (Fig.~\ref{fig:10K_synth_r20}), which illustrates that Mi3 is affected by out-of-sample error for typical dataset sizes, as previously described~\cite{haldane_influence_2019}. Indep has much lower $r_{20}$ scores than Mi3, as expected, since it does not model pairwise epistasis by design. Indep's $r_{20}$ scores are similar across all $r_{20}$ experiments, suggesting that it is not strongly affected by out-of-sample nor estimation error for this metric. This is expected, because its parameters are optimized for reproducing single-site frequency statistics only, which can be accurately estimated even from small training MSAs~\cite{haldane_influence_2019}. The $r_{20}$ scores for DeepSequence are slightly higher than sVAE's, but both VAEs remain close to each other for all training datasets and $n$, falling approximately halfway between Mi3 and Indep, and generally well below Mi3 at higher orders. Here in the 1M training MSA experiment, the VAEs' $r_{20}$ decreases to $\sim$0.4 and $\sim$0.5 at $n = 10$ for sVAE and DeepSequence respectively, reflecting higher specification error at higher orders (Fig.~\ref{fig:1M_synth_r20}). With 10K synthetic training sequences, the VAEs' $r_{20}$ decreases further for all $n$ due to the addition of out-of-sample error (Fig.~\ref{fig:10K_synth_r20}). 

Whereas the pairwise covariation correlations represent a preliminary indication that VAEs capture epistasis but mispredict its strength, the $r_{20}$ results reinforce this finding and extend it into higher orders. Unlike Mi3, the VAEs show specification error even when fit to large datasets from a model which only contains pairwise epistatic interaction terms (Fig.~\ref{fig:1M_synth_r20}). Because higher-order covariation statistics are constrained by the pairwise statistics, and the VAEs mispredict the pairwise statistics, we expect that the VAEs will exhibit specification error for higher-order epistasis, even though our synthetic test does not address this issue directly. When considered together with the $r_{20}$, the pairwise covariance correlations reveal a novel insight, which is that when trying to gauge GPSM generative capacity at higher orders of covariation, the standard pairwise statistics alone can be misleading. The relative magnitudes of $r_{20}$ between models at $n = 2$ are different at higher $n$, and the performance decrease as $n$ increases is more severe for the VAEs than Mi3 (Fig.~\ref{fig:covars_r20}, Bottom Row). \redb{In the natural analysis, Mi3 performs as close to the target as is measurable within error given the limited amount of data (Fig.~\ref{fig:10K_nat_r20}). Here, the small 10K target MSA causes large estimation error, making it more difficult to distinguish the performance of the different models. We emphasize that the $r_{20}$ performance decrease for Mi3 when tested against the synthetic and natural MSA targets can be accounted for entirely by out-of-sample error and estimation error, consistent with the hypothesis that pairwise interaction terms are sufficient to model protein sequence variation.}

\subsection*{Hamming distance distributions}

\begin{figure}
    \centering
    \begin{subfigure}[b]{0.3\textwidth}
        \centering
        \caption{}
        \label{fig:1M_synth_ham}
        \includegraphics[width=\textwidth]{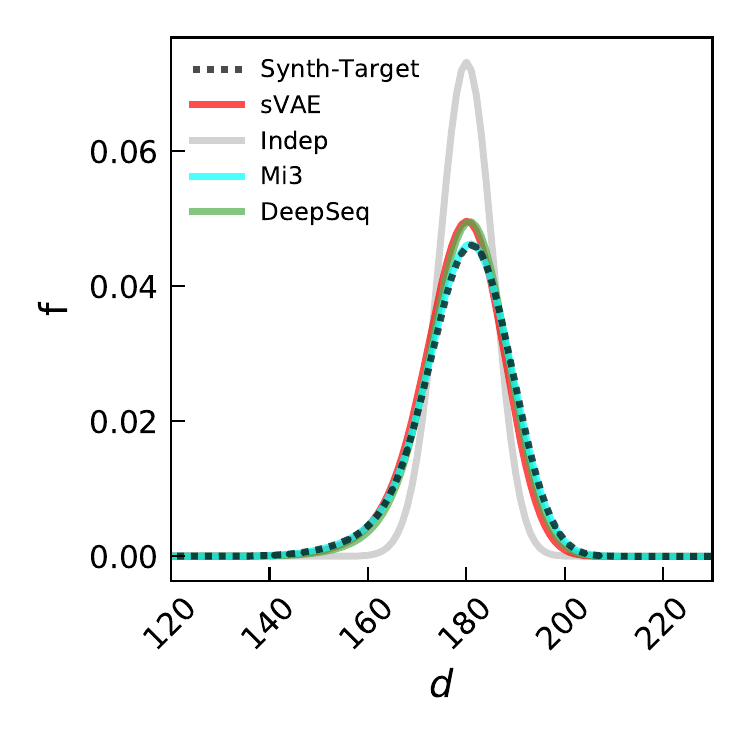}
    \end{subfigure}
    \hfill
    \begin{subfigure}[b]{0.3\textwidth}
        \centering
        \caption{}
        \label{fig:10K_synth_ham}
        \includegraphics[width=\textwidth]{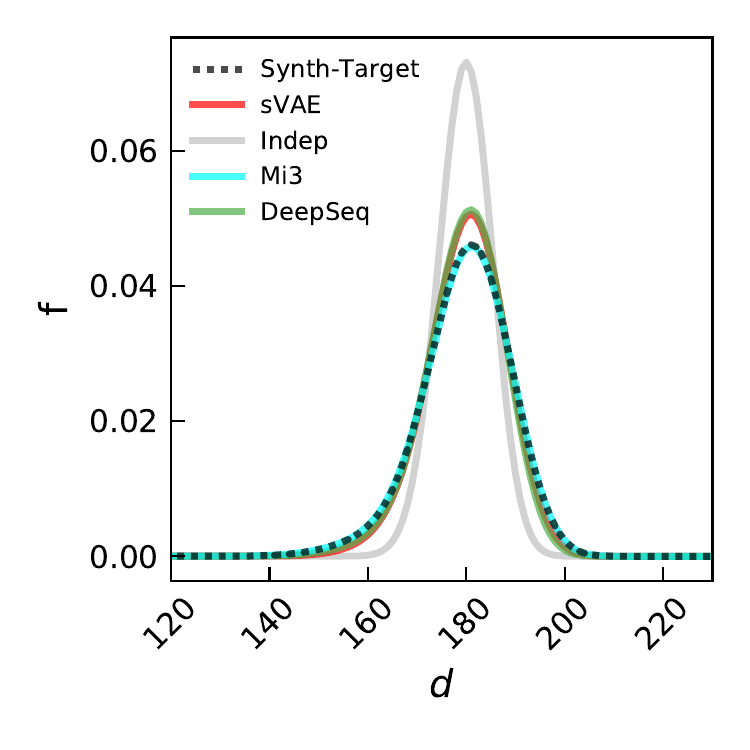}
    \end{subfigure}
    \hfill
    \begin{subfigure}[b]{0.3\textwidth}
        \centering
        \caption{}
        \label{fig:10K_nat_ham}        
        \includegraphics[width=\textwidth]{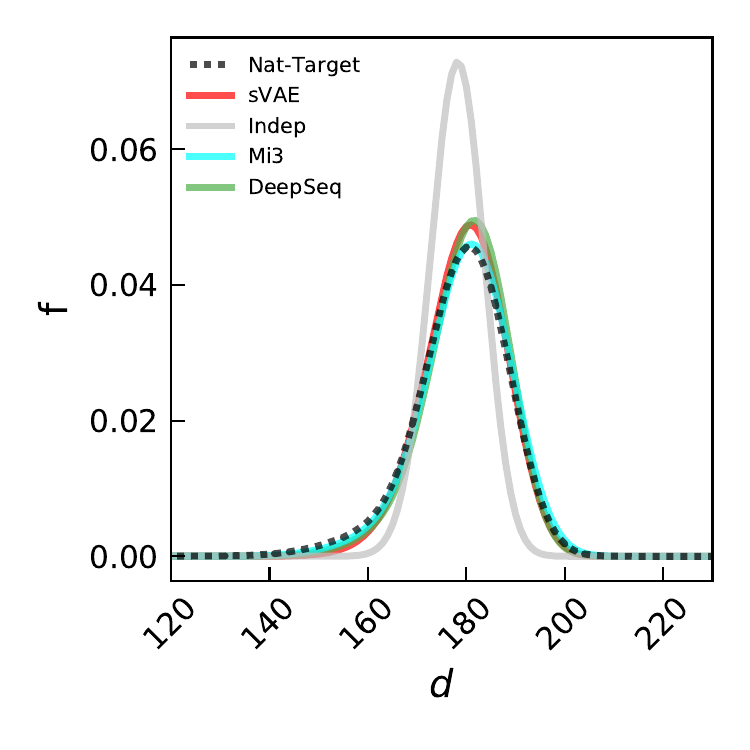}
    \end{subfigure}
    \hfill
    \begin{subfigure}[b]{0.3\textwidth}
        \centering
        \caption{}
        \label{fig:1M_synth_loglog}
        \includegraphics[width=\textwidth]{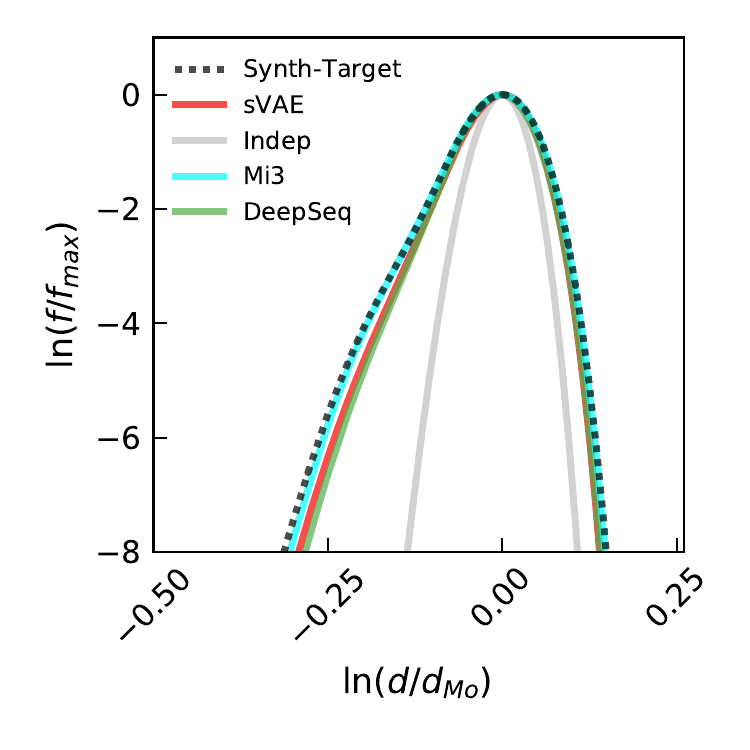}
    \end{subfigure}
    \hfill
    \begin{subfigure}[b]{0.3\textwidth}
        \centering
        \caption{}
        \label{fig:10K_synth_loglog}
        \includegraphics[width=\textwidth]{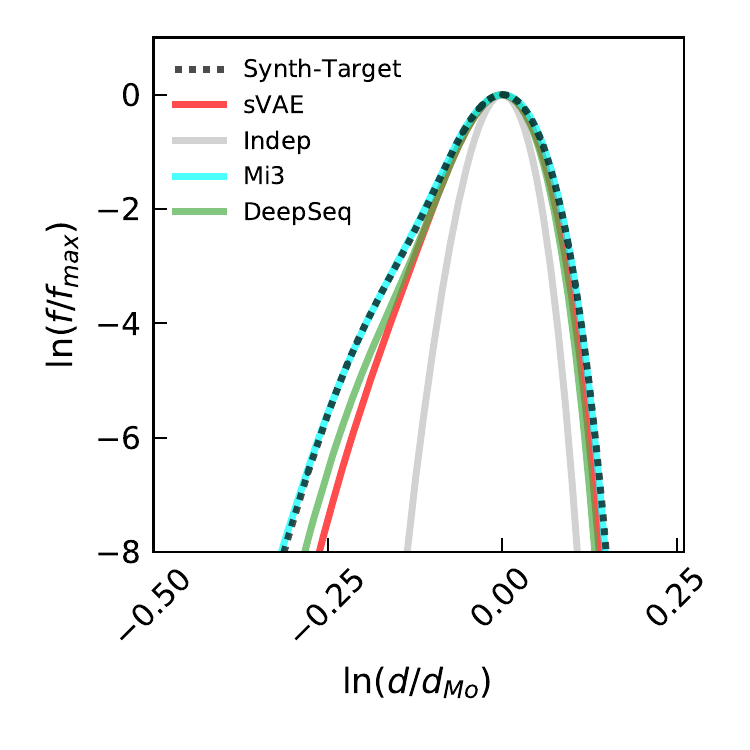}
    \end{subfigure}
    \hfill
    \begin{subfigure}[b]{0.3\textwidth}
        \centering
        \caption{}
        \label{fig:10K_nat_loglog}        
        \includegraphics[width=\textwidth]{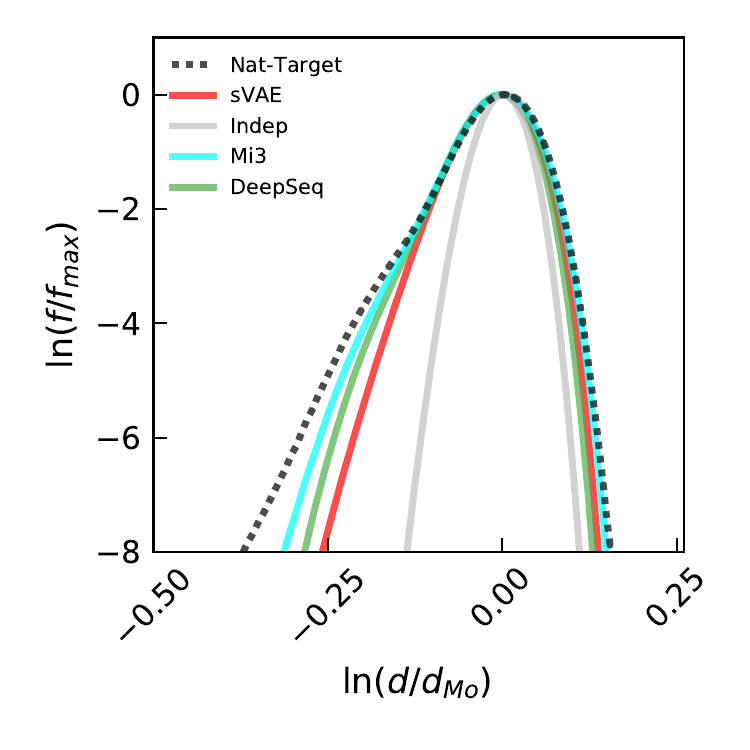}
    \end{subfigure}
	\caption{\textbf{Pairwise Hamming distance distributions.} These plots illustrate whether sequences generated from the GPSMs reproduce the overall sequence diversity of their respective targets. Mi3 (cyan), DeepSequence (green), sVAE (red), and Indep (gray) distributions are compared to target distribution (dotted black). GPSMs were trained on 1M synthetic (\textbf{a}, \textbf{d}), 10K synthetic (\textbf{b}, \textbf{e}), or 10K natural (\textbf{c}, \textbf{f}) sequences from the corresponding target distribution. All Hamming distributions were computed from 30K-sequence MSAs, except for the natural target, which was computed from a 10K-sequence target MSA due to data limitations. \textbf{Top Row}, Hamming distances $d$ (x-axis) are shown about the mode, and frequency $f$ is normalized as a fraction of total (y-axis). Mi3 perfectly matches the target distribution from \textbf{a} to \textbf{c}, whereas DeepSequence and sVAE overlap each other and share a mode with a slightly higher frequency than the target and Mi3. Indep has a slightly lower mode than all the other GPSMs, and a much higher mode frequency. Because generative capacity for all GPSMs is unchanged from \textbf{a} to \textbf{c}, none of them are sensitive to training or evaluation sample size for this metric. \textbf{Bottom Row}, Re-scaled logarithmic Hamming distance distributions better discriminate between GPSMs with respect to generative capacity than the normal Hamming distance distribution. Before being log-scaled, the Hamming distances $d$ are normalized by the mode $d_{Mo}$ (x-axis), and frequencies $f$ are normalized by the maximum Hamming distance $f_{max}$ (y-axis). This transformation highlights minute differences between distributions at low frequencies in the tails of the distributions on the left- and right-hand sides. From \textbf{d} to \textbf{e}, DeepSequence and sVAE appear sensitive to training sample size for this metric at the log-log scale.}
	\label{fig:hamming}
\end{figure}

We next evaluate the pairwise Hamming distance distribution metric $d(S, S')$. The Hamming distance between two protein sequences is the number of amino acids that are different between them, and we obtain a distribution for an MSA by comparing all pairs of sequences. Because it characterizes the range of sequence diversity in an MSA, recapitulation of the Hamming distance distribution has been used in the past as a measure for GPSM performance~\cite{cocco_inverse_2018,haldane_coevolutionary_2018,facco_intrinsic_2019,granata_accurate_2016}. In Fig.~\ref{fig:hamming}, we compare the pairwise Hamming distance distribution for each GPSM to that of the target distribution, computed with evaluation and target MSAs of 10K sequences each. To quantify the difference between the GPSM and target distributions for this metric, we use the total variation distance ($\operatorname{TVD}$)~\cite{levin_markov_2017}, which equals 1 when the distributions have no overlap and is 0 when they are identical, defined by $\operatorname{TVD}[f,g] = 1/2 \int |f(x) - g(x)| dx$.

All models reproduce the mode Hamming distance of $\sim$179. For Mi3, we report the same $\operatorname{TVD} = $0.007 when trained on either 1M (Fig.~\ref{fig:1M_synth_ham}) or 10K (Fig.~\ref{fig:10K_synth_ham}) synthetic sequences, showing negligible specification error, as expected. When trained on 10K natural sequences, Mi3 $\operatorname{TVD}$ increases to 0.012 (Fig.~\ref{fig:10K_nat_ham}), for reasons discussed further below. Indep \red{severely} underestimates the probability of both low and high Hamming distances, as observed at the distribution tails, with $\operatorname{TVD}\sim$0.24 across all experiments. The VAEs perform in between Mi3 and Indep, but much closer to Mi3 than Indep with respect to $\operatorname{TVD}$. Performance differences across all GPSMs for this metric indicate that out-of-sample error has a consistent and detectable, though very small, effect on the fundamental sequence diversity of artificial GPSM-generated MSAs. That Mi3 and the VAEs are highly performant and comparable to each other, but not Indep, corroborates our earlier findings that epistasis is relevant to accurate modelling of protein sequence diversity (Fig.~\ref{fig:covars_r20}). However, because Indep performs much closer to Mi3 and the VAEs for this metric than on any other, and also because this metric cannot discriminate well between Mi3 and the VAEs, we suspect that reproducing the Hamming distance distribution is a much easier hurdle for GPSMs than is reproducing higher-order covariation. This shortcoming of the standard Hamming distance distribution metric becomes apparent when these results are compared to those of our novel metric, $r_{20}$, which does show a significant gap in generative capacity between Mi3 and the VAEs at higher orders (Fig.~\ref{fig:covars_r20}, Bottom Row).

To emphasize the decay of the tails, we rescale all the distributions by their maxima and re-center them around their modes to give them the same peak, and then plot them on a log-log scale (Fig.~\ref{fig:hamming}, Bottom Row). The relevance of the distributions' tails lies in their power-law behavior as they approach zero, where the function's exponent is related to the intrinsic dimension of the dataset and therefore to the number of informative latent factors needed to explain the data~\cite{facco_intrinsic_2019,granata_accurate_2016,ansuini2019}. A well-specified GPSM ought to reproduce this exponent, and therefore the tail's decay, since it is a topological property intrinsic to the dataset and independent from the particular choice of variables used to describe the probability density~\cite{ansuini2019}. There is a trend of slightly decreasing generative capacity as training samples decrease, which is detectable only here in the log-scaled Hamming distribution (Fig.~\ref{fig:hamming}, Bottom Row). In this modified rendering of the Hamming distance distribution, differences in GPSM generative capacity can be observed at both low (Left Tail) and high (Right Tail) sequence diversity. The Mi3 distribution closely overlaps the target distribution with both 1M (Fig.~\ref{fig:1M_synth_loglog}) and 10K (Fig.~\ref{fig:10K_synth_loglog}) synthetic training sequences. In the 10K natural experiment, Mi3 deviates noticeably from the target on the left tail (Fig.~\ref{fig:10K_nat_loglog}), which represents less evolutionarily diverged sequences. This could be an artifact of the phylogenetic relationships between sequences present in the natural dataset, which may have been incompletely removed by our phylogenetic filtering step for this dataset (see Methods, Supplementary Information), or it could be due to estimation error in measuring the target distribution, as only 10K target sequences are available to estimate the black line in the natural analysis. As before, Indep performance is consistently low across all experiments. The VAEs' performance at low sequence diversity (Fig.~\ref{fig:hamming}, Bottom Row, Left Tails) decreases for smaller training dataset size.

\subsection*{Statistical Energy Correlations}

A fourth metric we use to evaluate generative capacity is the statistical energy $E(S)$ of individual sequences in the dataset, which we express using the negative logarithm of the predicted sequence probability $p(S)$, where $E(S) = -\log p(S)$. $E(S)$ can be computed analytically for Mi3 and Indep, and estimated for VAE models by importance sampling (see Methods).

\begin{figure}
    \centering
    \begin{subfigure}[b]{0.26\textwidth}
        \centering
        \caption{}
        \label{fig:1M_mi3_energy}
        \includegraphics[width=\textwidth]{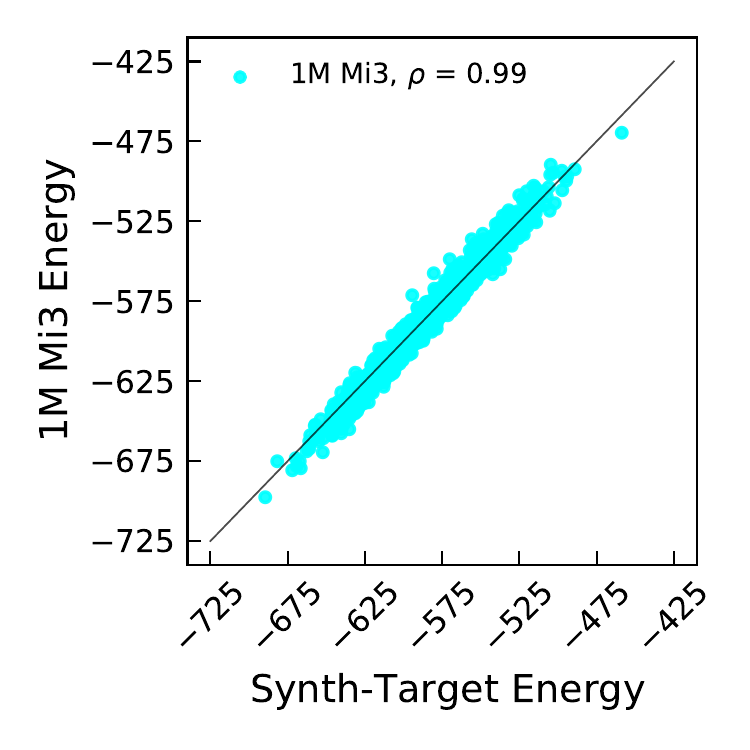}
    \end{subfigure}
    \begin{subfigure}[b]{0.26\textwidth}
        \centering
        \caption{}
        \label{fig:10K_mi3_energy}
        \includegraphics[width=\textwidth]{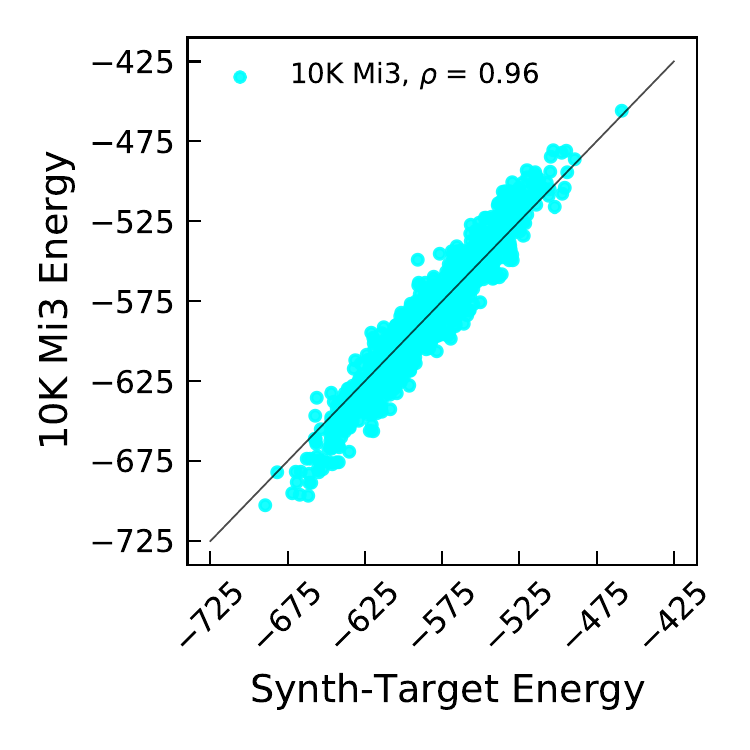}
    \end{subfigure}
    \\
    \begin{subfigure}[b]{0.26\textwidth}
        \centering
        \caption{}
        \label{fig:1M_DS_energy}        
        \includegraphics[width=\textwidth]{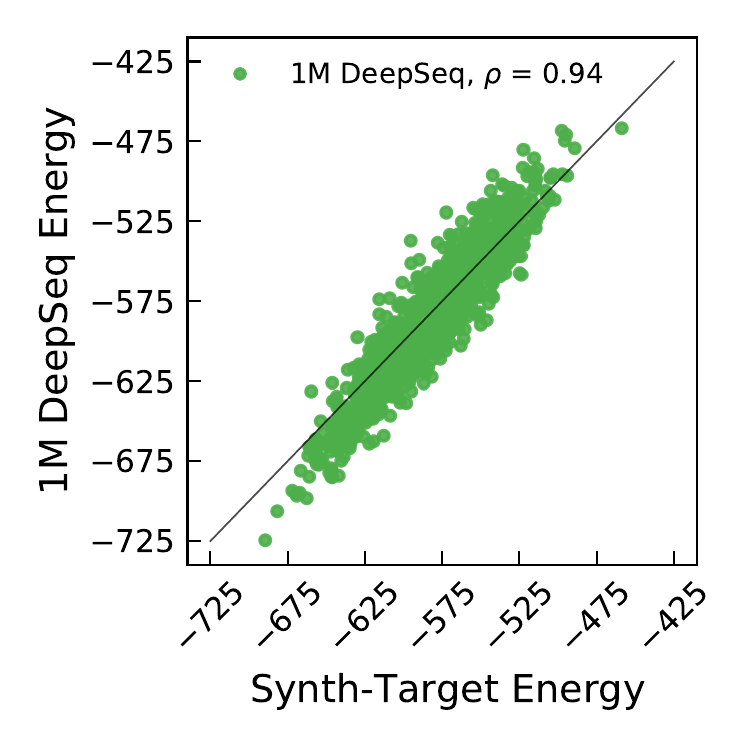}
    \end{subfigure}
    \begin{subfigure}[b]{0.26\textwidth}
        \centering
        \caption{}
        \label{fig:10K_DS_energy}
        \includegraphics[width=\textwidth]{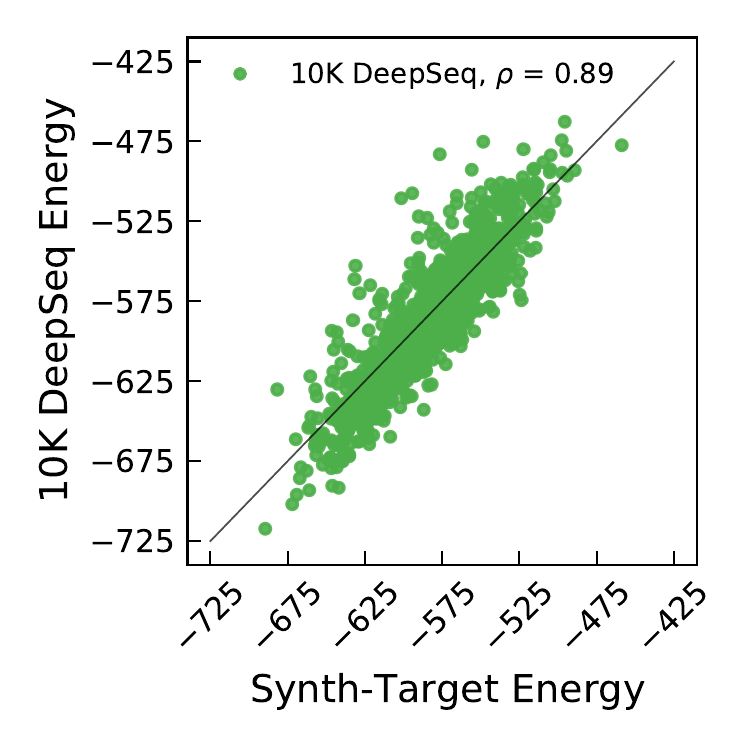}
    \end{subfigure}
    \\
    \begin{subfigure}[b]{0.26\textwidth}
        \centering
        \caption{}
        \label{fig:1M_vae_energy}
        \includegraphics[width=\textwidth]{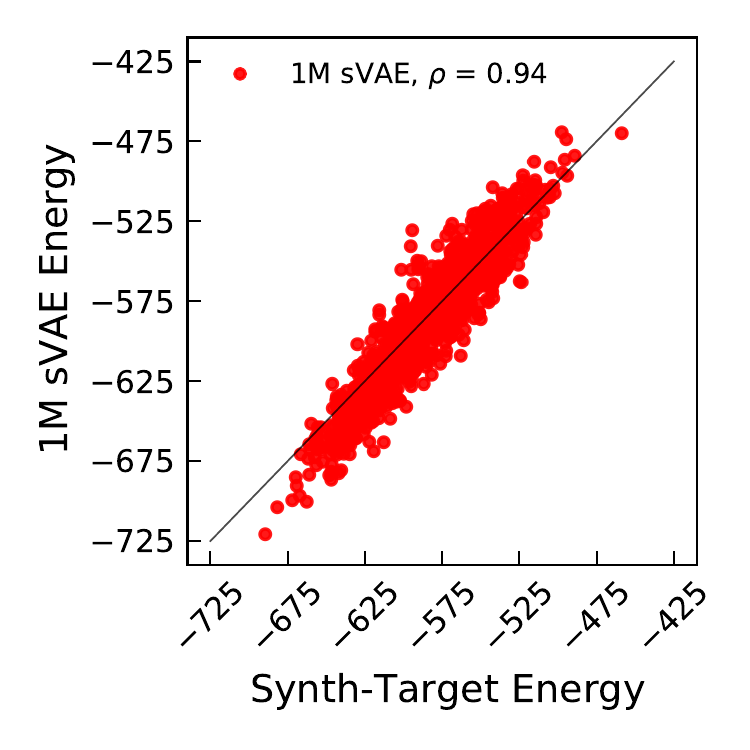}
    \end{subfigure}
    \begin{subfigure}[b]{0.26\textwidth}
        \centering
        \caption{}
        \label{fig:10K_vae_energy}        
        \includegraphics[width=\textwidth]{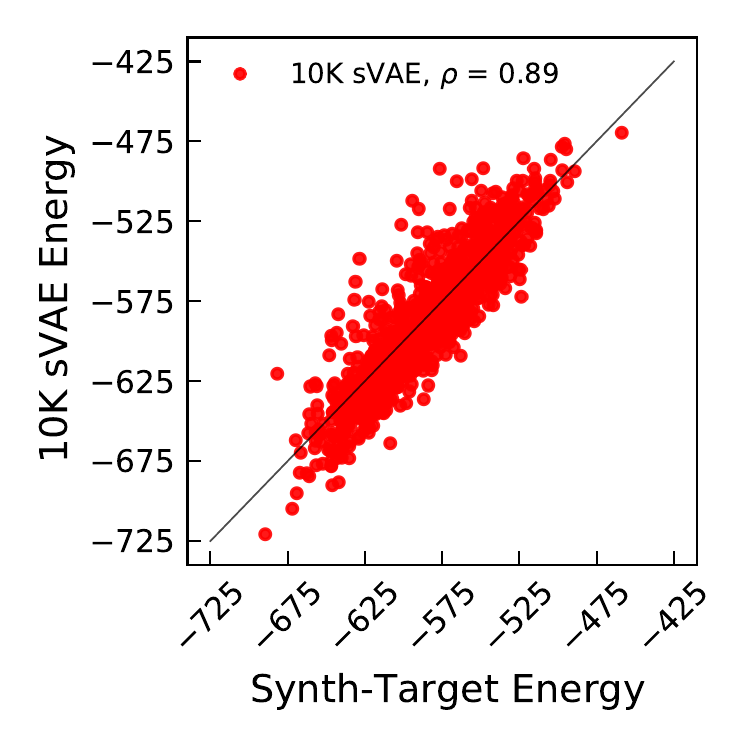}
    \end{subfigure}
    \\
    \begin{subfigure}[b]{0.26\textwidth}
        \centering
        \caption{}
        \label{fig:1M_indep_energy}
        \includegraphics[width=\textwidth]{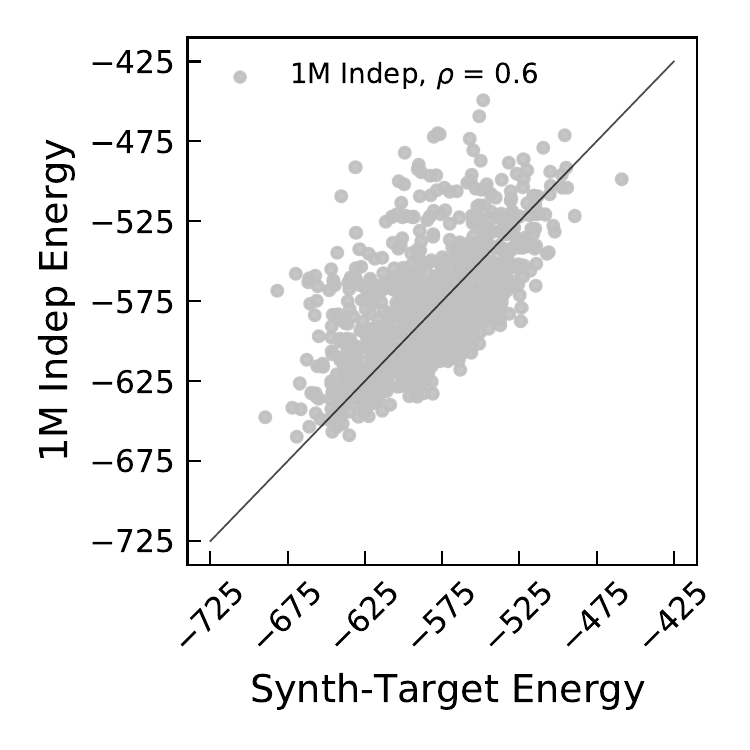}
    \end{subfigure}
    \begin{subfigure}[b]{0.26\textwidth}
        \centering
        \caption{}
        \label{fig:10K_indep_energy}        
        \includegraphics[width=\textwidth]{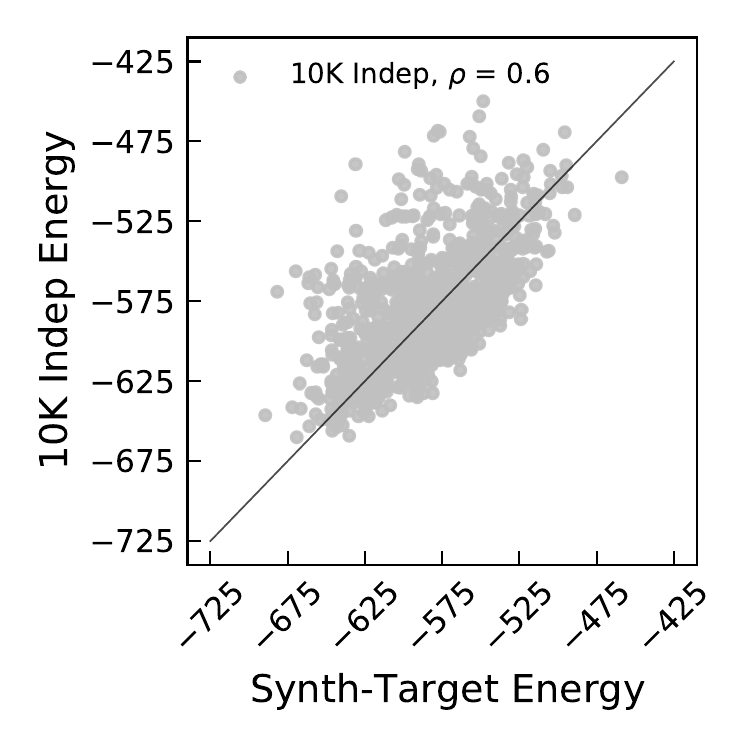}
    \end{subfigure}
	\caption{\textbf{Statistical energy correlations.} Statistical energies $E(S)$ of 1K synthetic test sequences from the target distribution as evaluated by Mi3 (cyan), DeepSequence (green), sVAE (red), Indep (gray). Each GPSM was trained on 1M (\textbf{a}, \textbf{c}, \textbf{e}, \textbf{g}) or 10K (\textbf{b}, \textbf{d}, \textbf{f}, \textbf{h}) sequences from the synthetic target distribution. For each scatterplot, Pearson correlation coefficient $\rho$ was computed between each GPSM’s statistical energy and that of the synthetic target distribution for each sequence. These statistical energies are unit-less, and so we have shifted the data along the y-axis to compare $E(S)$ on the same scale. Only Indep is insensitive to decreased training sample size for this metric.}
	\label{fig:energies}
\end{figure}

This statistic directly evaluates accuracy of the GPSM distribution values from $p_\theta(S)$ for a limited number of individual sequences, which has been used to validate GPSMs by comparison to corresponding experimental fitness values~\cite{riesselman_accelerating_2019,riesselman_deep_2018,figliuzzi_coevolutionary_2016,haldane_coevolutionary_2018}. In Fig.~\ref{fig:energies}, we compare artificial statistical energies from the GPSM distribution $p_\theta(S)$ to those of the target distribution $p^0(S)$ for a 1K test MSA generated from $p^0(S)$. \redb{This measurement cannot be performed for the natural 10K experiment because $p^0(S)$ is unknown for the natural data, so we present results only for the 1M and 10K synthetic experiments.} We use the 1M (Fig.~\ref{fig:energies}, Left Column) and 10K (Fig.~\ref{fig:energies}, Right Column) synthetic training MSA sizes, and quantify GPSM generative capacity for this metric by the Pearson correlation coefficient $\rho(\{E(S)\}, \{\hat{E}(S)\})$ between synthetic target energies $E(S)$ and GPSM energies $\hat{E}(S)$. Mi3 reproduces the synthetic target distribution at both training MSA sizes. Because Mi3 should have very low specification error on the synthetic target, as it is well specified by design, the small amount of error must be due to remaining out-of-sample or numerical errors. As expected, Indep poorly reproduces the target values, with $\rho = 0.6$ for both MSA training sizes. The VAEs exhibit slightly larger specification error than Mi3 on the 1M training set with correlation of $\rho = 0.94$, and exhibit further out-of-sample error on the 10K training set with $\rho = 0.89$. 

\redb{Juxtaposing the $E(S)$ results to the $r_{20}$ results reiterates the striking insight of our work, which is that despite the utility of standard metrics for measuring sequence statistics, they say little about a GPSM's ability to capture higher-order covariation, and therefore necessarily higher-order epistasis. If considered by itself, the $E(S)$ metric would indicate that both VAEs have generative capacity close to that of Mi3, and even Indep could be said to have a large amount of generative capacity, in spite of capturing no covariation by design. But according to $r_{20}$, which directly measures the higher-order marginals, the VAEs and Indep are comparable to Mi3 only at the pairwise level. Critically, at higher orders, the performance difference between the VAEs to each other remains relatively unchanged, but is significantly lower than Mi3, with Indep lower still. This suggests that $E(S)$, just as with the Hamming distance metric and pairwise covariance correlation, represents an easier, and perhaps different, hurdle for GPSMs than measuring generative capacity by the ability to capture higher-order sequence covariation, as done uniquely by $r_{20}$.}

\section*{Conclusions}

\redb{In this study we reveal the steep challenges and limits entailed by current measurements of generative capacity in GPSMs trained on either synthetic or currently available natural sequence datasets.} Recent state-of-the-art GPSM studies have benchmarked their models by comparing $p_\theta(S)$ to experimental fitness values from deep mutational scans~\cite{riesselman_deep_2018,sinai_variational_2018}, or by generating artificial sequences that appear to fold into realistic structures based on \textit{in silico} folding energy~\cite{madani_progen_2020}. However, these strategies for model evaluation present their own challenges. Point mutation fitness experiments in practice may measure particular contributors to fitness, including replicative capacity, drug resistance, protein stability, and enzymatic activity~\cite{riesselman_deep_2018}, and are subject to significant experimental error and other limitations, e.g. those imposed by conservation~\cite{haddox_mapping_2018}. Likewise, \textit{in silico} computational chemistry approaches rely on energy functionals that may lack the precision needed to meaningfully discriminate between highly similar sequences~\cite{alley_unified_2019}. Additionally, neither protein function nor fitness rely exclusively on the thermodynamic stability of static native structures, but also on the protein's conformational dynamics~\cite{wei_protein_2016,anishchenko_origins_2017,sailer_molecular_2017,petrovic_conformational_2018,nussinov_protein_2019}, which are not fully described by folding energy values alone~\cite{campitelli_role_2020}. This could mean that despite generating sequences with realistic \textit{in silico} folding energy, a GPSM may still not be capturing crucial higher-order epistatic effects. Neither point mutation fitness effects, nor \textit{in silico} folding energy estimations, are directly related to mutational covariation statistics observed in an MSA in the sense that they do not check if subsets of covarying amino acids in specific positions present in the target MSA are indeed present in the GPSM-generated evaluation MSA. Our novel $r_{20}$ metric uniquely delivers that functionality, emphasizing higher-order covariation where previous studies rarely go beyond the pairwise level~\cite{cocco_inverse_2018,haldane_coevolutionary_2018,haq_pairwise_2009,shimagaki_selection_2019,hawkins-hooker_generating_2021}.

Benchmarking coevolution-based protein sequence models in data rich and data poor regimes, as done here, is an effective method for ascertaining where data-driven effects stop, and algorithmic failure begins~\cite{alquraishi_proteinnet_2019}. Due to the limited availability of natural protein sequence data, this line is inherently blurred in the natural analysis, as we demonstrate across all four of our generative capacity metrics. But in our synthetic analysis track, we have demonstrated the extent to which VAEs, with different implementations, can capture higher-order covariation at orders between three and ten when the target distribution is known, its statistical properties are measurable with a high degree of certainty, and major forms of error are removed, minimized, or accounted for. When given a large number of training and target sequences, we found both VAEs' generative capacity to be between that of a site-independent model (Indep) and a pairwise Hamiltonian (Mi3) for all measurements. In the synthetic $r_{20}$ tests, our results show that both VAEs' generative capacity is well below Mi3, raising questions about whether VAEs can capture higher-order epistasis significantly better than a pairwise Potts model. \red{In our synthetic analysis the target distribution is a Potts model, and therefore we expect Mi3 to fit the target distribution well by design. However, we find Mi3 also outperforms the VAEs where we do not have this expectation, such as on the natural target distribution and on a target distribution specified by sVAE, as shown in Supplementary Information, suggesting Mi3 generally outperforms VAEs on protein sequence data with respect to generative capacity.}

The Hamming distance distributions, pairwise covariation correlations, and statistical energy correlations are standard metrics that have been used in the past to measure GPSM accuracy, but we find that they can be inadequate or misleading indicators of a GPSM’s ability to capture covariation at higher orders. Taken together, our results suggest that, of the metrics we tested, only $r_{20}$ provides the granularity needed to discriminate between different GPSM’s ability to model higher-order epistasis, as it directly tests the model's ability to capture higher-order covariation. 

Although our results suggest VAEs are less effective for capturing higher-order epistasis than pairwise Potts models, VAEs have demonstrated utility in unsupervised learning and clustering. One VAE-GPSM, ``BioSeqVAE'', has generated artificial sequences that share a ``hallucinated'' homology to natural proteins in the training set, which could mean that their folded structures would perform similar functions to their hallucinated natural homologs~\cite{costello_how_2019}. Another, ``PEVAE'', has shown that a VAE-GPSM’s latent space captures phylogenetic relationships~\cite{ding_deciphering_2019} better than PCA~\cite{bishop_pattern_2006} and t-SNE~\cite{maaten_visualizing_2008}. These VAE-GPSMs furnished a latent space that immediately allowed for function-based protein classification, a benefit unavailable to pairwise Potts models without some effort.

\redb{The causes of VAE performance limitations are actively being investigated in literature from multiple directions. One line of inquiry involves a VAE phenomenon known as ``posterior collapse'' (see Supporting information), in which some dimensions of the VAE latent space become insensitive to the input data. Studies of this phenomenon have led to some insights into VAE behavior, for instance that in some situations the VAE likelihood can contain spurious local maxima\cite{lucas2019understanding}, and many different heuristic strategies to understand and avoid this phenomenon have been suggested~\cite{dai2020the,fu_cyclical_2019,takahashi_variational_2019}. In Supporting Information we test that sVAE, used in our main results, does not exhibit posterior collapse, though we can trigger it for sVAE architectures with more than 7 latent dimensions in the bottleneck layer. Another line of inquiry relates to assumptions typically made about the metric and topology of the latent space, suggesting that the commonly used Euclidean metric space or Gaussian prior distribution may not best describe particular datasets, for instance because of an effect called ``manifold mismatch''~\cite{davidson_hyperspherical_2018,falorsi_explorations_2018}. Techniques closely related to the VAE, such as the ``WAE'', are also being investigated as alternatives to the VAE, with different performance characteristics\cite{tolstikhin2018wasserstein}. Among these competing frameworks for exploring VAE behavior, we have tested VAE implementations which are currently used to generatively model protein sequences. With more nuanced latent variable models, and with better understanding of protein sequence embeddings, perhaps GPSM generative capacity could extend beyond what has been demonstrated here with state-of-the-art VAEs.}

Our \red{epistasis-oriented methodology} focuses on measuring higher-order covariation, with the potential for broad applicability to various sequentially ordered data. \red{$r_{20}$-like} measurements become possible when the data are sufficient in number, and the correlation structures between elements, both within and across samples, are statistically detectable and meaningful in some context, be it visual, biophysical, or linguistic. The convergence between data categories such as images, proteins, and language with respect to generative modelling evaluation offers the exciting opportunity of a wider, interdisciplinary audience for the work proposed here. \red{Conversely, further development of sophisticated, data-intensive, and direct generative capacity metrics of GPSMs could reveal nuances of the correlation structure of protein sequence datasets that distinguish them from other datasets, helping to explain why $r_{20}$-like metrics can detect higher-order covariation, whereas other metrics cannot.} \redb{Our work represents not only a revision of currently prevailing paradigms of GPSM benchmarking, but also a challenge to generative protein sequence modelling more broadly, to consider how epistasis and direct higher-order covariation metrics like $r_{20}$ can inform their models and results.}

\section*{Methods}

\subsection*{Sequence dataset preparation}

\redb{An outline of our sequence processing is shown in Fig.\ref{fig:pipeline}. In Stage 1 we obtain sequences.} For the natural analysis, we use an MSA of the kinase protein superfamily which we have previously curated using sequences from the Uniprot/TREMBL database~\cite{the_uniprot_consortium_uniprot_2019}. This MSA is composed of $\sim$20K sequences of length 232, obtained by filtering a larger set of $\sim$291K sequences to remove any sequences with more than 50\% sequence identity to another (see Supplementary Information). \redb{For the synthetic analysis, we treat the Potts Hamiltonian model trained on this natural protein kinase MSA as the target model or distribution, and generate MSAs from it which serve as training and target MSAs in our synthetic tests.

In Stage 2 we split the sequences into training and target MSAs. For the natural analysis, we randomly divide the 20K natural sequences into 10K training and 10K target MSAs. For the synthetic analysis we generate both training MSAs used to train the synthetic models, and target MSAs used as reference, and ensure that the synthetic training and target sequences are non-overlapping sets, even though they come from the same target model.

In Stage 3, the processed MSAs are then one-hot encoded and used to train the the GPSMs. In Stage 4, we generate evaluation MSAs from each model, and in Stage 5 we perform our generative capacity measurements by comparing the artificial MSAs to the appropriate target MSA.}

\subsection*{Mi3} 

The Mi3 model is a pairwise Potts Hamiltonian model fit to sequence data using the ``Mi3-GPU'' software we have developed previously~\cite{haldane_mi3-gpu_2020}, which performs ``inverse Ising inference'' to infer parameters of Potts models using a Markov-Chain Monte-Carlo (MCMC) algorithm which entails very few approximations. This software allows us to fit statistically accurate Potts models to MSA data. We have examined Mi3's generative capacity and out-of-sample error in earlier work~\cite{haldane_mi3-gpu_2020,haldane_influence_2019}, which we summarize here.

A Potts model is the maximum entropy model for $p(S)$ constrained to reproduce the bivariate marginals $f^{ij}_{\alpha\beta}$ of an MSA, i.e. the frequency of amino acid combination $\alpha,\beta$ ,at positions $i,j$. The probability distribution $p_\theta(S)$ for the Potts model takes the form
\begin{equation}
    p_\theta(S) = \frac{e^{-E(S)}}{Z} \quad \text{with} \quad E(S) = \sum_i^L h^i_{s_i} + \sum_{i<j} J^{ij}_{s_i s_j}
    \label{Potts_P}
\end{equation}
where $Z$ is a normalization constant, $Z = \sum_S e^{-E(S)}$, and ``coupling'' $J^{ij}_{\alpha\beta}$ and ``field'' $h^i_\alpha$ parameters are compactly denoted by the vector $\theta = \{h^i_\alpha, J^{ij}_{\alpha\beta}\}$. The number of free parameters of the model (non-independent couplings and fields) is equal to the number of non-independent bivariate marginals, which can be shown to be $\frac{L(L-1)}{2}(q-1)^2+L(q-1)$ for $q$ amino acids~\cite{bialek_rediscovering_2007}, which is $\sim$10.7M parameters for our model. This implies that the Potts model is well specified to reproduce the bivariate marginals when generating sequences from $p_\theta(S)$. 

The Mi3 model inference procedure maximizes the log-likelihood with regularization. Maximizing the Potts log-likelihood can be shown to be equivalent to minimizing the difference between the dataset MSA bivariate marginals $f^{ij}_{\alpha\beta}$ and the model bivariate marginals of sequences generated from $p(S)$. To account for finite sampling error in the estimate of $f^{ij}_{\alpha\beta}$ for an MSA of $N$ sequences we add a small pseudocount of size $1/N$, as described previously~\cite{haldane_influence_2019}. We also add a regularization penalty to the likelihood affecting the coupling parameters $J^{ij}_{\alpha\beta}$ to bias them towards 0, of form $\lambda \sum \operatorname{SCAD}(J^{ij}_{\alpha\beta}, \lambda, \alpha)$ using the SCAD function which behaves like $\lambda |J^{ij}_{\alpha\beta}|$ for small $J^{ij}_{\alpha\beta}$ but gives no bias for large $J^{ij}_{\alpha\beta}$\cite{SCAD}, using a small regularization strength of $\lambda=0.001$ for all inferences which causes little model bias.

To generate synthetic MSAs from the Potts model we use MCMC over the trial distribution $p_\theta(S)$ until the Markov-Chains reach equilibrium, as described previously~\cite{haldane_mi3-gpu_2020}. We can directly evaluate $E(S)$ as the negative log-probability of any sequences for the Mi3 model using equation \ref{Potts_P} up to a constant $Z$, and this constant can be dropped without affecting our results.

\subsection*{Indep}

The Indep model is the maximum entropy model for $p(S)$ constrained to reproduce the univariate marginals of an MSA, and is commonly called a ``site-independent'' model because the sequence variations at each site are independent of the variation at other sites. Because it does not fit the bivariate marginals, it cannot model covariation between positions. It takes the form
\begin{equation}
    p_\theta(S) = \frac{e^{-E(S)}}{Z} \quad \text{with} \quad E(S) = \sum_i^L h^i_{s_i} 
    \label{Indep_P}
\end{equation}
where $Z$ is a normalization constant, $Z = \sum_S e^{-E(S)}$, and ``field'' parameters $h^i_\alpha$ for all positions $i$ and amino acids $\alpha$ are compactly referred to by the vector $\theta = \{h^i_\alpha\}$. The fields of the Indep model generally have different values from the fields of the Potts model. Unlike for the Potts model, maximum likelihood parameters can be determined analytically to be $h^i_\alpha = -\log f^i_\alpha$ where $f^i_\alpha$ are the univariate marginals of the dataset MSA. When fitting the Indep model to a dataset MSA of $N$ sequences, we add a pseudocount of $1/N$ to the univariate marginals to give model marginals $\hat{f}^i_\alpha$ , to account for finite sample error in the univariate marginal estimates. The model distribution simplifies to a product over positions, as $p_\theta(S)=\sum \hat{f}^i_{s_i}$. The number of independent field parameters is $L(q-1)$ which equals ~4.6K parameters for our model.

To generate sequences from the independent model we independently generate the residues at each position $i$ by a weighted random sample from the marginals $f^i_\alpha$, and we directly evaluate the log probability of each sequence $E(S)$ from equation \ref{Indep_P}.

\subsection*{VAEs}  

The standard variational autoencoder (sVAE) is a deep, symmetrical, and undercomplete autoencoder neural network composed of a separate encoder $q_\phi(Z|S)$ and decoder $p_\theta(S|Z)$~\cite{charte_practical_2018}, which map input sequences $S$ to regions within a low-dimensional latent space $Z$ and back. The probability distribution for the sVAE is defined as
\begin{equation}
    p_\theta(S) = \int p_\theta(S|Z) p(Z) dZ
    \label{VAE_P}
\end{equation}
where the latent space distribution is a unit Normal distribution, $p(Z) = \mathcal{N}[ 0,1](Z)$. Training of a VAE can be understood as maximization of the dataset log-likelihood with the addition of a Kullback-Leibler regularization term $\operatorname{D_{KL}}[q_\phi(Z|S), p_\theta(Z|S)]$, where $p_\theta(Z|S)$ is the posterior of the decoder~\cite{kingma_auto-encoding_2014,rezende_stochastic_2014}.

\redb{sVAE's architecture is ``vanilla''~\cite{ding_guided_2020}, meaning it is implemented in a standard way and its behavior is meant to be representative of VAEs generally. Nearly identical sVAE architectures have been used as VAE-GPSMs in recent studies~\cite{sinai_variational_2018}. sVAE's encoder and decoder are implemented without advancements such as convolutional layers~\cite{riesselman_deep_2018}, multi-stage training~\cite{dai_diagnosing_2019}, disentanglement learning~\cite{ding_guided_2020}, Riemannian Brownian motion priors~\cite{kalatzis_variational_2020}, and more. This allows us to directly interrogate the assumptions and performance of standard variational autoencoding with respect to the training and evaluation of GPSMs in this work.}

Our encoder and decoder have 3 layers each and employ standard normalization and regularization strategies~\cite{ioffe_batch_2015,srivastava_dropout_2014}. sVAE's latent bottleneck layer has 7 nodes, and the model in total has ~2.7M inferred parameters. The input layer of the encoder accepts a one-hot encoded sequence and the decoder's output layer values can be interpreted as a Bernoulli distribution of the same dimensions as a one-hot encoded sequence. We have tested various sVAE architectures and hyperparameters with our datasets, as well as DeepSequence \redb{as described below}, and found qualitatively similar generative capacity results.

To generate a sequence from sVAE, we generate a random sample in latent space from the latent distribution $p(Z)$ pass this value to the decoder to obtain a Bernoulli distribution, from which we sample once. To evaluate the negative log-probability of a sequence $E(S)$ we use importance sampling, averaging over 1K samples from the latent distribution $q_\phi(Z|S)$~\cite{ding_deciphering_2019}. Other publications use the Evidence Lower Bound (ELBO) estimate as an approximation of the negative log-probability~\cite{riesselman_deep_2018}, and we have verified that the ELBO and the negative log-probability are nearly identical in our tests and have equal computational complexity.

\redb{In addition to sVAE, we test the DeepSequence VAE~\cite{riesselman_deep_2018}. We use the default inference parameters, and use the ``SVI'' inference implementation which uses a ``variational Bayes'' inference technique as an extension of the sVAE inference method. Because DeepSequence is designed to output the ELBO rather than the negative log-probability for each sequence, we use the ELBO as an approximation of $E(S)$, and estimate it using an average over 1K samples. To generate sequences we use the same strategy as for sVAE.}

\bibliography{ms}

\section*{Acknowledgements}

This research was supported by the National Science Foundation, grant number NSF-GCR 1934848, and the National Institute of Health, grant number R35-GM132090. This research includes calculations carried out on HPC resources supported in part by the National Science Foundation through major research instrumentation grant number 1625061 and by the US Army Research Laboratory under contract number W911NF-16-2-0189.

\section*{Author contributions statement}

F.M., R.M.L, V.C., A.H. conceived the experiments. F.M., V.C., A.H. performed the experiments. F.M., R.M.L, V.C., A.H. analyzed the results. F.M., V.C., A.H. wrote the bulk of the codebase, Q.N. made a contribution to the codebase. F.M., R.M.L, V.C., A.H. wrote the paper. All authors reviewed the manuscript. 

\end{document}

\begin{center}
    \Large
    \textbf{Supplementary Information for:} 
    
    \textbf{Generative Capacity of Probabilistic Protein Sequence Models}
    
    \vspace{0.4cm}
    \normalsize
    \text{Francisco McGee \qquad Quentin Novinger \qquad Ronald M. Levy \qquad Vincenzo Carnevale \qquad Allan Haldane}
    \vspace{0.9cm}
\end{center}

\section{sVAE implementation}

The standard variational autoencoder (sVAE) is a deep, symmetrical, and undercomplete autoencoder neural network composed of a separate encoder $q_\phi(Z|S)$ and decoder $p_\theta(S|Z)$, which map input sequences $S$ to regions of a low-dimensional latent space $Z$ and back.\cite{kingma_auto-encoding_2014} It is a probabilistic model, and in our ``vanilla''~\cite{ding_guided_2020} implementation we assume sequences will be distributed according to a unit normal distribution in latent space, $p(Z) = \mathcal{N}[0,1](Z)$\cite{kingma_auto-encoding_2014}. Training of a VAE can be understood as maximization of (the logarithm of) the dataset likelihood $\mathcal{L} = \sum_S p_\theta(S) = \sum_S \int p_\theta(S|Z) p(Z) dZ $ with the addition of a Kullback-Leibler regularization term $\operatorname{D_{KL}}[q_\phi(Z|S), p_\theta(Z|S)]$, where $p_\theta(Z|S)$ is the posterior of the decoder, which allows use of the fitted encoder $q_\phi(Z|S)$ to perform efficient estimation of the likelihood and its gradient by Monte-Carlo sampling, for appropriate encoder models.
 
Our sVAE architecture is built on the same basic VAE architecture of ``EVOVAE''\cite{sinai_variational_2018}, which itself appears to be built on the VAE implementation provided by developers for the Keras library\cite{chollet_keras_2015}. It is composed of 3 symmetrical ELU-activated layers in both the encoder and decoder, each layer with 250 dense (fully-connected) nodes. The encoder and decoder are connected by a latent layer of $l$ nodes, we use $l=7$ in the main text. sVAE's input layer accepts one-hot encoded sequences, the output layer is sigmoid-activated, and its node output values can be interpreted as a Bernoulli distribution of the same dimensions as a one-hot encoded sequence. The first layer of the encoder and the middle layer of the decoder have dropout regularization applied with $30\%$ dropout rate, and the middle layer of the encoder uses batch normalization with a batch size of 200\cite{srivastava_dropout_2014, ioffe_batch_2015,sinai_variational_2018}. In all inferences, we hold out 10\% of the training sequences as a validation dataset, and perform maximum likelihood optimization using the Keras Adam stochastic gradient optimizer on the remaining 90\%\cite{kingma_adam_2017}. After each training epoch we evaluate the loss function for the training and validation data subsets separately. We have tested using early-stopping regularization to stop inference once the validation loss has not decreased for three epochs in a row, as in previous implementations, but this led to some variability in the model depending on when the early stopping criterion was reached. To avoid this variability, and to make different models more directly comparable, we instead fixed the number of epochs to 32 for all models, since in the early stopping tests this led to near minimum training loss and validation loss, and did not lead to significant overfitting as would be apparent from an increase in the validation loss.

sVAE was implemented using Keras, building on previous implementations\cite{sinai_variational_2018,chollet_keras_2015}, however with a modification of the loss function relative to both of these, to remove a scaling factor of $Lq$ on the reconstruction loss, which is sometimes used to avoid issues with local minima as described further below. This prefactor leads to a non-unit variance of the latent space distribution of the dataset sequences, violating our definition that the latent space distribution should be normal with unit variance, $p(Z) = \mathcal{N}[0,1](Z)$. In the next section we show that after removing the prefactor the latent space distribution is approximately a unit normal, which more closely follows the original VAE conception\cite{kingma_auto-encoding_2014,rezende_stochastic_2014}. Our implementation is available at \url{https://github.com/ahaldane/MSA_VAE}.

To generate a sequence from the model we generate a random sample in latent space from the latent distribution $\mathcal{N}[0,1]$, pass this value to the decoder to obtain a Bernoulli distribution, from which we sample once. To evaluate the log-probability of a sequence, we use importance sampling, averaging over 1000 samples from the latent distribution $q_\phi(Z|S)$ following from the relations  \cite{MAL-056,ding_deciphering_2019}
\begin{equation}
\begin{split}
    p_\theta(S) &= \int p_\theta(S|Z) p(Z) dZ = \int q_\phi(Z|S) \frac{p_\theta(S|Z)p(Z)}{q_\phi(Z|S)} dZ \\
   &= \underset{Z \sim q_\phi(Z|S)}{\mathbb{E}} \left[  \frac{p_\theta(S|Z)p(Z)}{q_\phi(Z|S)}  \right] \approx \frac{1}{N} \sum_i^N \frac{p_\theta(S|Z^i)p(Z^i)}{q_\phi(Z^i|S)}
\end{split}
\end{equation}
where, $Z^i$ are independent samples from $q_\phi(Z|S)$ and $N$ is a large number of samples. Here $q_\theta(Z|S)$ plays the role of a sampling bias function, biasing samples to regions of latent space which are likely to have generated the sequence, leading to an accurate Monte-Carlo estimate of $p_\theta(S)$. The value $p_\theta(S)$ can be converted to a unit-less statistical energy as $E(S) = -\log p_\theta(S)$ for direct comparison with Mi3 and Indep statistical energies. 

\begin{figure*}
\centering
\includegraphics[width=3.4in]{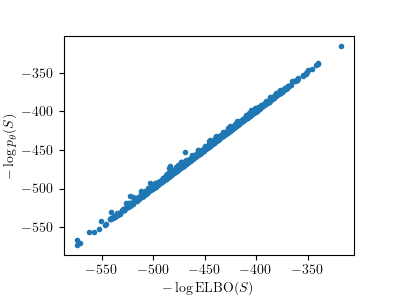}
\caption{Comparison of $E(S) = -\log p_\theta(S)$ with the ELBO estimate for the sVAE with $l=7$ fit to 1M sequences, evaluated for 1000 sequences $S$ from the validation dataset, with $N=1000$ samples for both the ELBO estimate and the $E(S)$ estimate.}
\label{fig:ELBO}
\end{figure*}

Other publications have used the Evidence Lower Bound (ELBO) estimate as an approximation of $\log p_\theta(S)$\cite{riesselman_deep_2018}, and we have tested (see Fig. \ref{fig:ELBO}) that the ELBO and the log-probability are nearly identical, and $N=1000$ samples is sufficient for an accurate estimate. The fact that the ELBO and log-probability are nearly identical is a sign that our encoder is well fit, as the difference between these values should equal the KL divergence  $\operatorname{D_{KL}}[q_\phi(Z|S), p_\theta(Z|S)]$ between the ``true'' posterior of the decoder $p_\theta(Z|S)$ and the approximate posterior $q_\phi(Z|S)$, which should be 0 if the encoder $q_\phi(Z|S)$ has accurately modelled the posterior\cite{kingma_auto-encoding_2014}.

\section{VAE model validation and generalization} 

\begin{figure}
    \centering
    \begin{subfigure}[b]{0.3\textwidth}
        \centering
        \caption{}
        \label{fig:l2}
        \includegraphics[width=\textwidth]{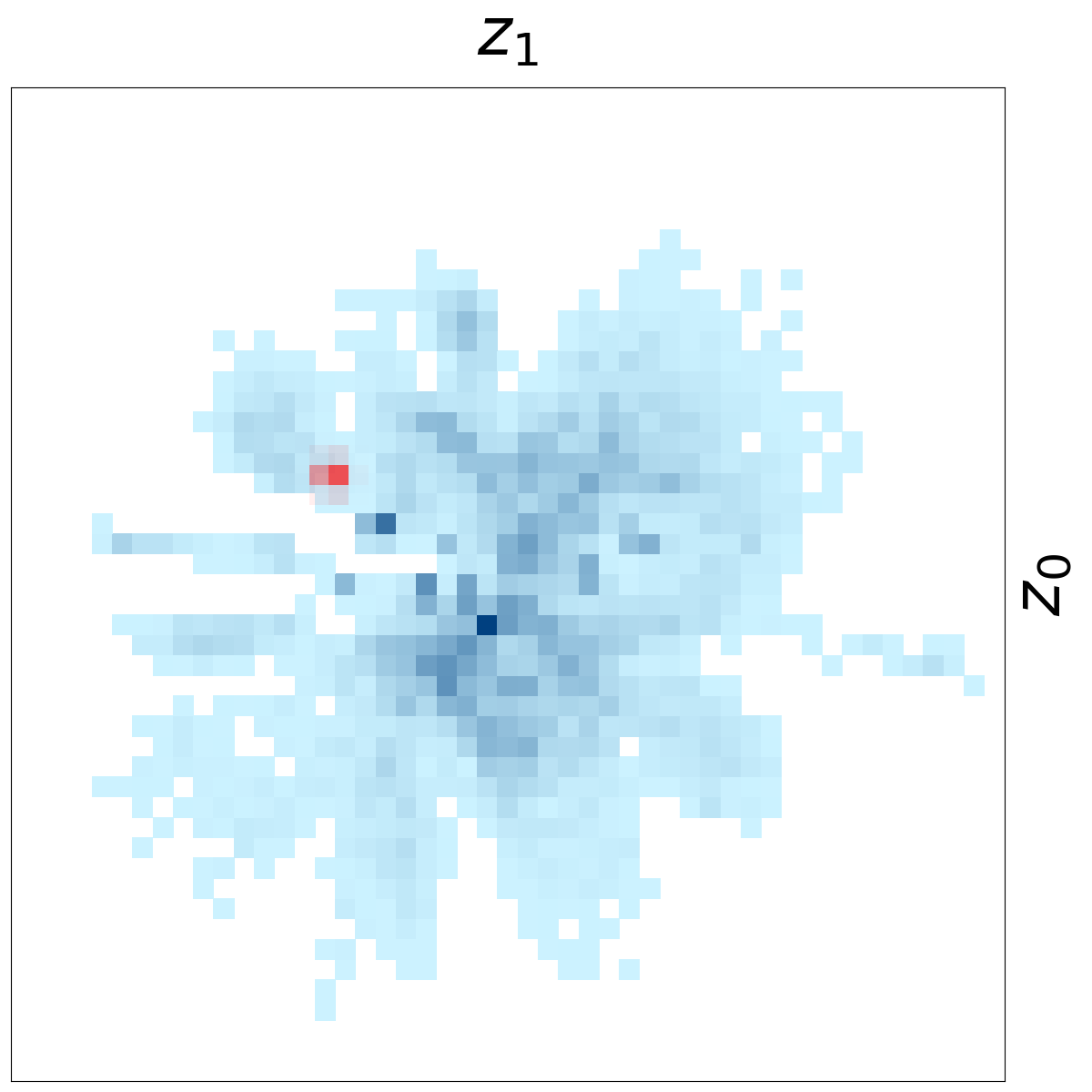}
    \end{subfigure}
    \begin{subfigure}[b]{0.3\textwidth}
        \centering
        \caption{}
        \label{fig:l4}
        \includegraphics[width=\textwidth]{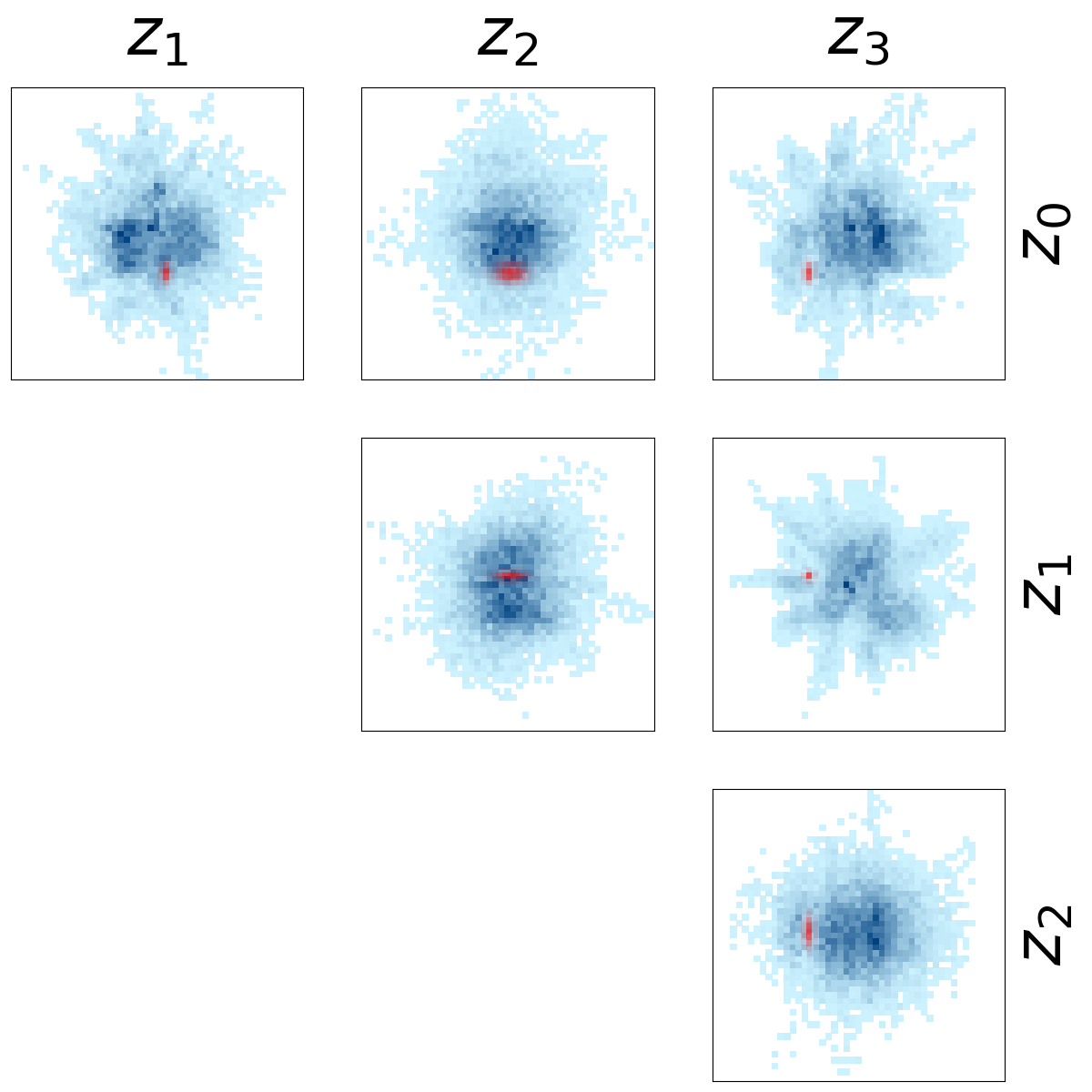}
    \end{subfigure}
    \begin{subfigure}[b]{0.3\textwidth}
        \centering
        \caption{}
        \label{fig:l6}        
        \includegraphics[width=\textwidth]{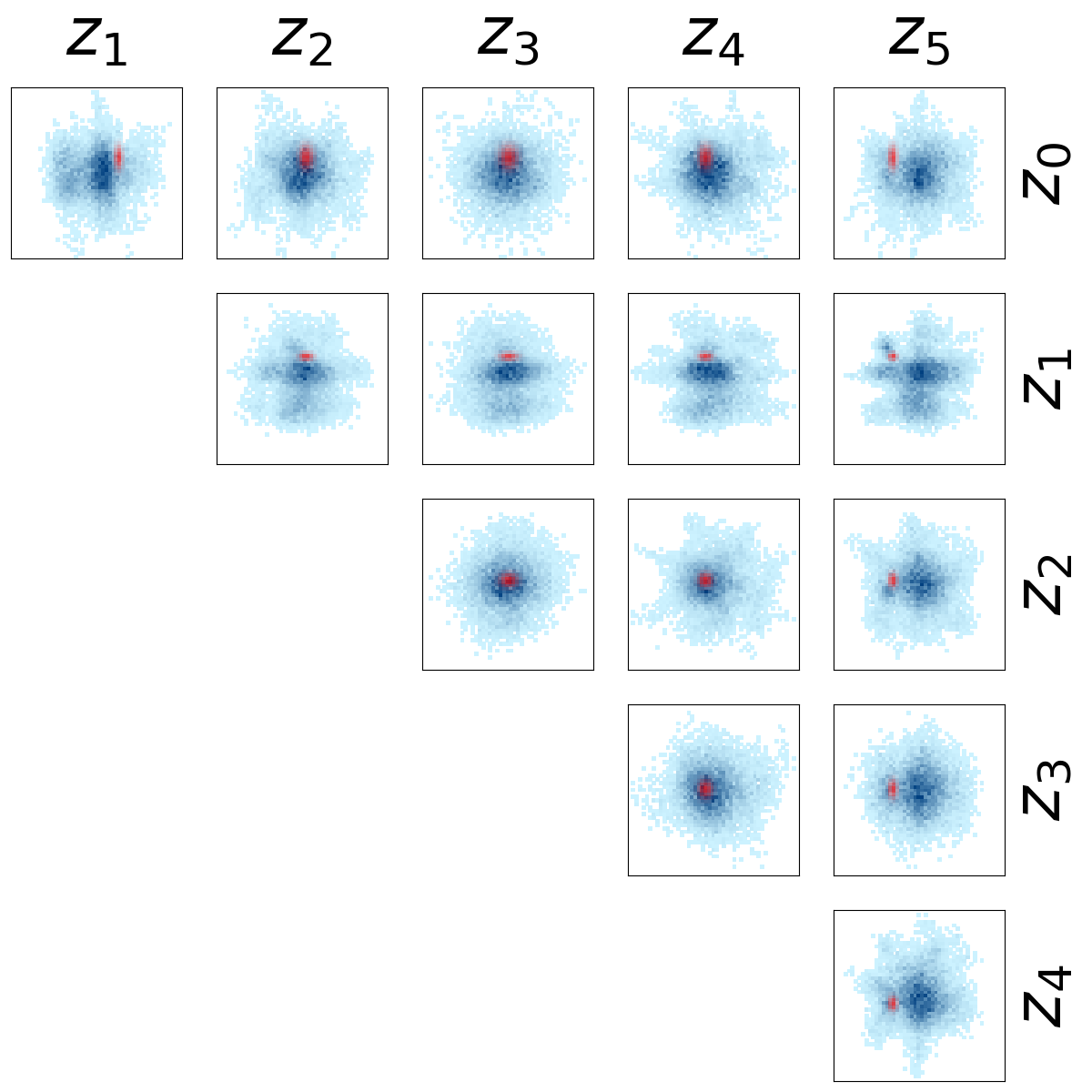}
    \end{subfigure}
    \begin{subfigure}[b]{0.45\textwidth}
        \centering
        \caption{}
        \label{fig:l8}
        \includegraphics[width=\textwidth]{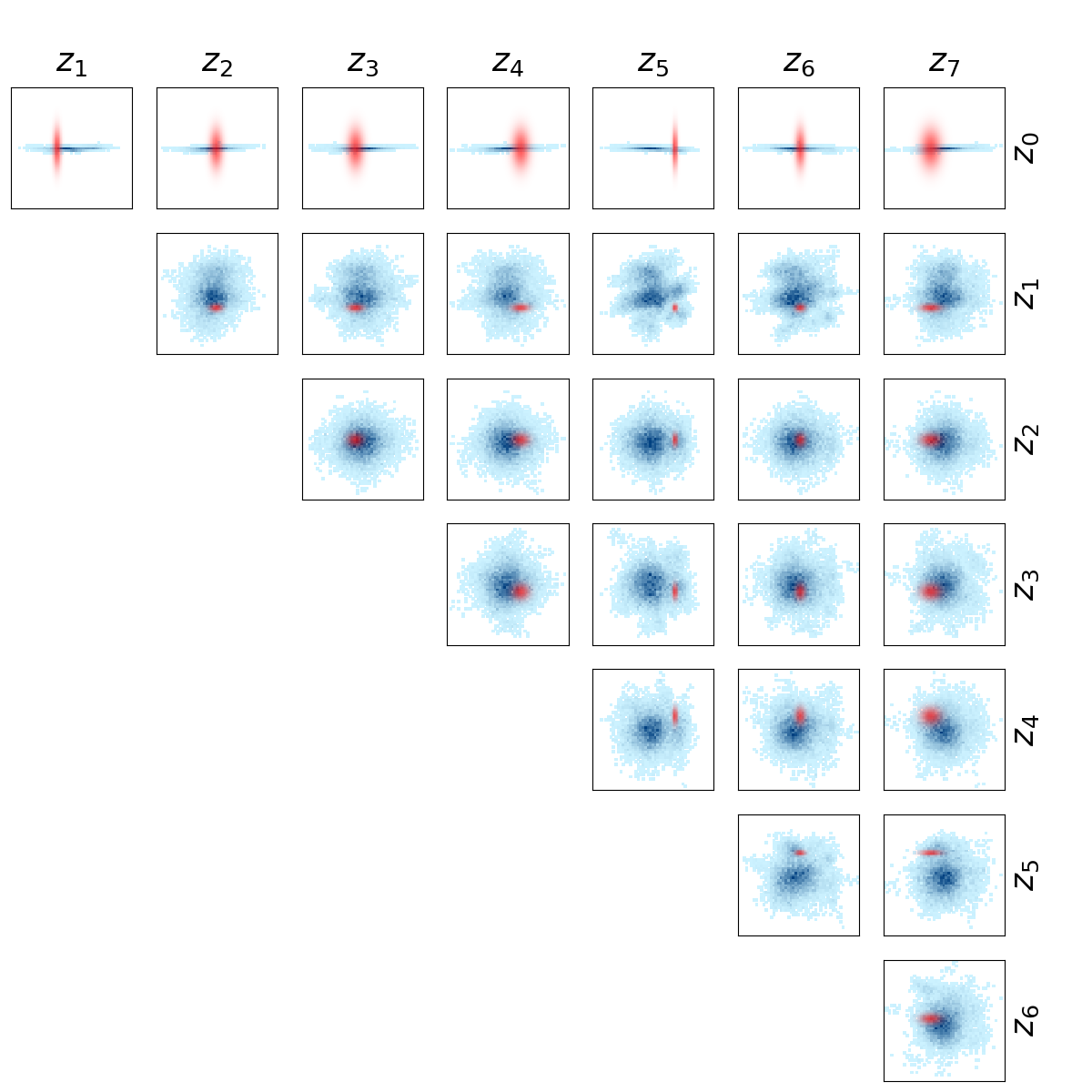}
    \end{subfigure}
    \begin{subfigure}[b]{0.45\textwidth}
        \centering
        \caption{}
        \label{fig:l10}
        \includegraphics[width=\textwidth]{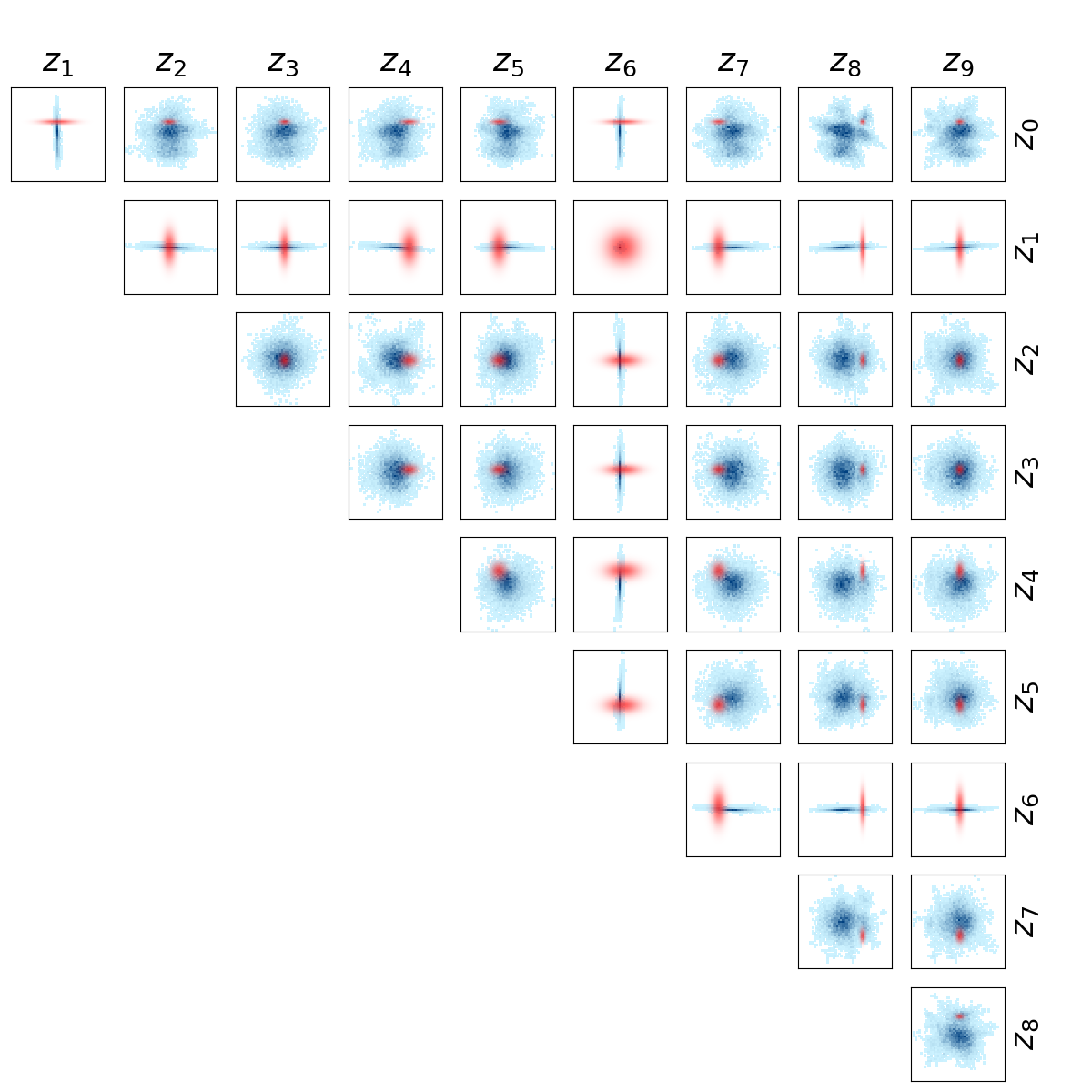}
    \end{subfigure}
	\caption{Plots of latent space distribution of the training dataset for sVAE models fit with different latent space sizes of 2, 4, 6, 8, and 10 (\textbf{a},\textbf{b},\textbf{c},\textbf{d},\textbf{e} respectively), fit to 1M synthetic training sequences as in the synthetic test in the main text. For each latent space size we show, for each pair of latent variables, a 2d histogram of the projected means of 10K training dataset sequences in latent space in blue. There is one subplot for $l=2$, six subplots for $l=4$, etc. Each plot ranges from -4 to 4 on both axes. The latent distribution $q_\phi(Z|S)$ for single random sequence from the training dataset is shown as a red shading in proportion to probability.}
	\label{fig:latent_training}
\end{figure}

To validate our choice of latent space size of $l=7$ used in the main text, we tried fitting sVAEs with different latent space sizes from 2 to 10. In Fig. \ref{fig:latent_training} we illustrate the latent space projections of the sequences in the training dataset. According to our specification underlying the VAE theoretically, we expect the latent space distribution to be a multidimensional normal distribution with mean 0 and unit variance. Indeed, as can be seen in the plot, and measured numerically, we generally find the latent space distribution of the dataset has close to unit variance and is approximately normal, although there is some non-normal structure in the distribution.

For latent spaces of $l=8$ and $l=10$ we observe that some latent dimensions appear to have``collapsed'', in particular $z_0$ for $l=8$, and $z_1$ and $z_6$ for $l=10$. From repeated runs (not shown) we observe that the number of collapsed dimensions varies somewhat depending on the random seed used to initialize the stochastic optimizer, and also depends on the size of the training dataset as more dimensions collapse when fitting 1M sequences than fitting 10K sequences (not shown). For these ``collapsed'' dimensions, we see that the projected variance of the illustrated sequence in red in Fig. \ref{fig:latent_training} is very close to 1.0, unlike in other dimensions where the projected variance is much smaller. These behaviors are consistent with a well known phenomenon of ``posterior collapse'' discussed in VAE literature\cite{lucas2019understanding}. It has been suggested that VAE posterior collapse can occur due to local minima in the likelihood function which are not global minima\cite{lucas2019understanding}, but in some situations can be a sign that additional latent dimensions are uninformative, and that fewer latent dimensions better represent the data\cite{dai2020the}. We find that choosing $l=7$ gives the best performing model which avoids posterior collapse. Interestingly, the number of ``informative'' latent variables, i.e. those that do not undergo posterior collapse, turns out to coincide with the intrinsic dimension (ID) of the dataset of training sequences, estimated from the set of pairwise distances using a completely independent approach~\cite{granata_accurate_2016}. In brief, it has been shown that graph distances calculated on k-neighbor graphs can be used to approximate geodesics and thus to generate the distribution of ``intrinsic'' distances. Close to the maximum, the latter depends exclusively on the dimension of the distance distribution's support. This observation is used to devise a family of estimators for the ID. Using these tools, we estimated an ID of 7 or 8 for the synthetic dataset used in the main text. These numbers are consistent with what was observed in terms of collapse of the posterior distribution: the ID is seemingly related to the number of informative latent variables so that if the number of nodes in the embedding layer is increased past this number, then posterior collapse occurs, indicating that the additional variables are not needed to explain the data.            

\begin{figure}
    \centering
    \begin{subfigure}[b]{0.45\textwidth}
        \centering
        \caption{}
        \label{fig:latent_r20_10K}
        \includegraphics[width=\textwidth]{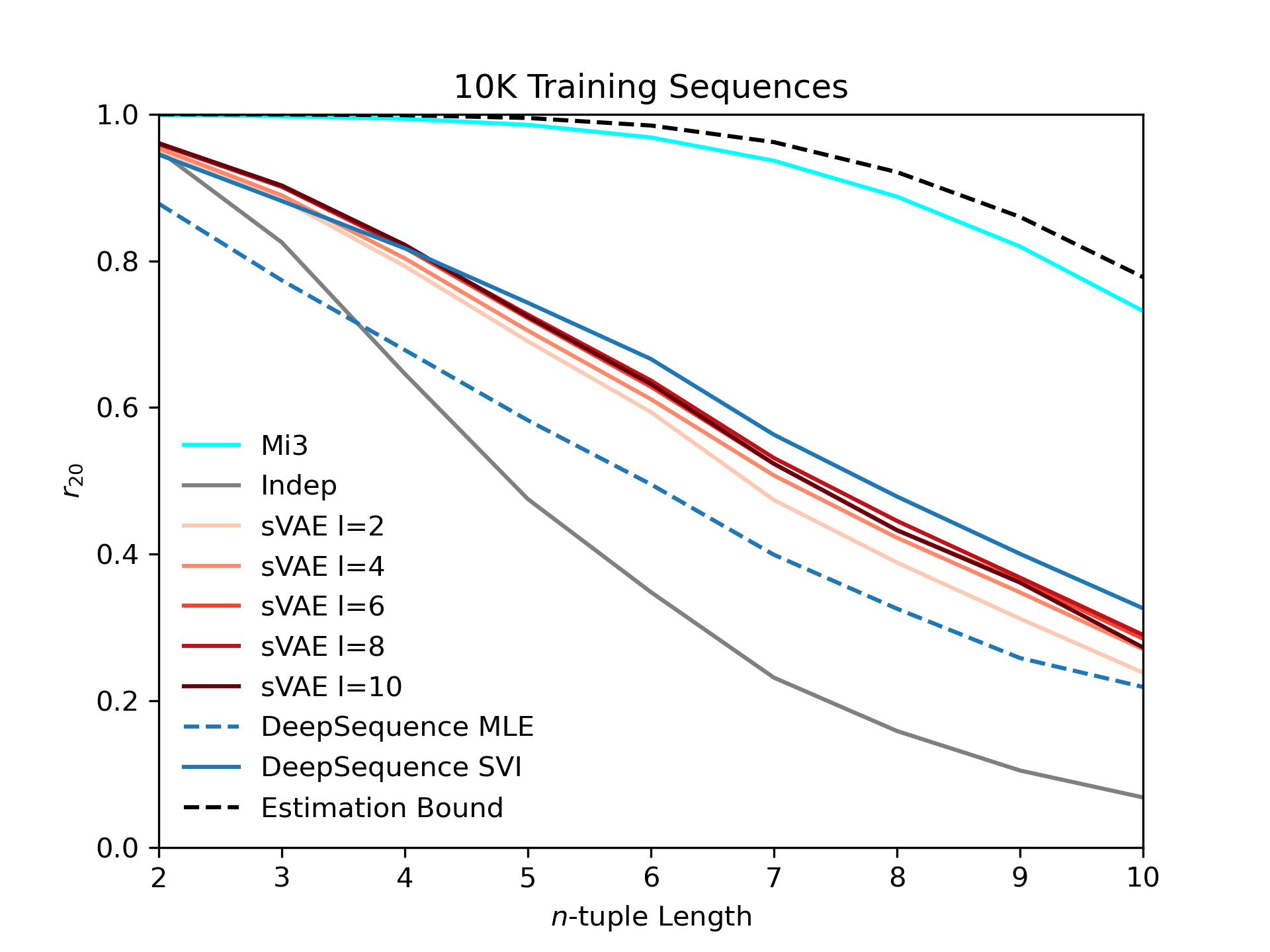}
    \end{subfigure}
    \begin{subfigure}[b]{0.45\textwidth}
        \centering
        \caption{}
        \label{fig:latent_r20_1M}
        \includegraphics[width=\textwidth]{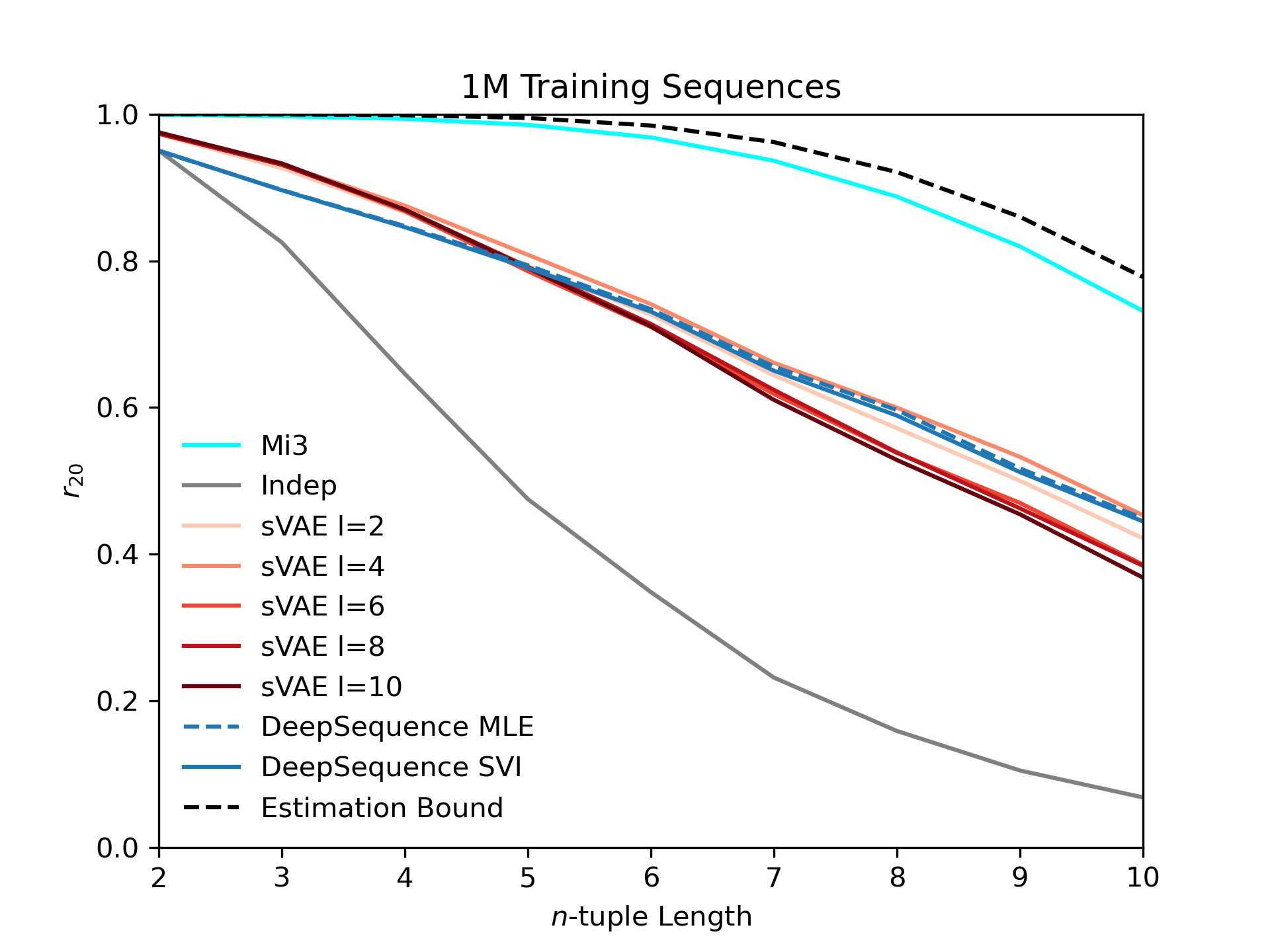}
    \end{subfigure}
	\caption{Performance comparison of sVAEs for different $l$ compared to DeepSequence VAEs and Mi3 using the $r_{20}$ metric on 10K synthetic (\textbf{a}) and 1M synthetic (\textbf{b}) training sequences.}
\label{fig:latent_r20}
\end{figure}

To compare the generative capacity of different GPSMs to determine how general our results are, we computed our MSA statistics for other VAEs besides the $l=7$ sVAE shown in the main text. In Fig. \ref{fig:latent_r20} we show the $r_{20}$ scores for different models when fit to either 10K or 1M synthetic sequences, as in the synthetic tests in the main text. We include the Mi3 and Indep models, as well as sVAEs for different latent space sizes, and also models produced using the DeepSequence VAE software which comes in two variations, the ``MLE'' and the ``SVI'' algorithms\cite{riesselman_deep_2018}, for which we use the default or example parameters. All the VAEs perform fairly similarly in this metric, including the DeepSequence VAEs. For the smaller training dataset of 10K sequence the DeepSequence SVI algorithm outperforms the other VAEs, suggesting it is less susceptible to out-of-sample error. These results suggest that our results for the sVAE shown in the main text generalize to other VAEs, including the significantly more complex DeepSequence VAE, and are not strongly dependent on implementation or number of latent variables $l$. The models with $l\sim 7$ perform among the best of the sVAE models for both the 10K and the 1M training datasets, though the difference between the models is small, and this further justifies our choice of $l=7$ in the main text.

\section{Minimizing VAE out-of-sample error}

\begin{figure*}
\centering
\includegraphics[width=3.4in]{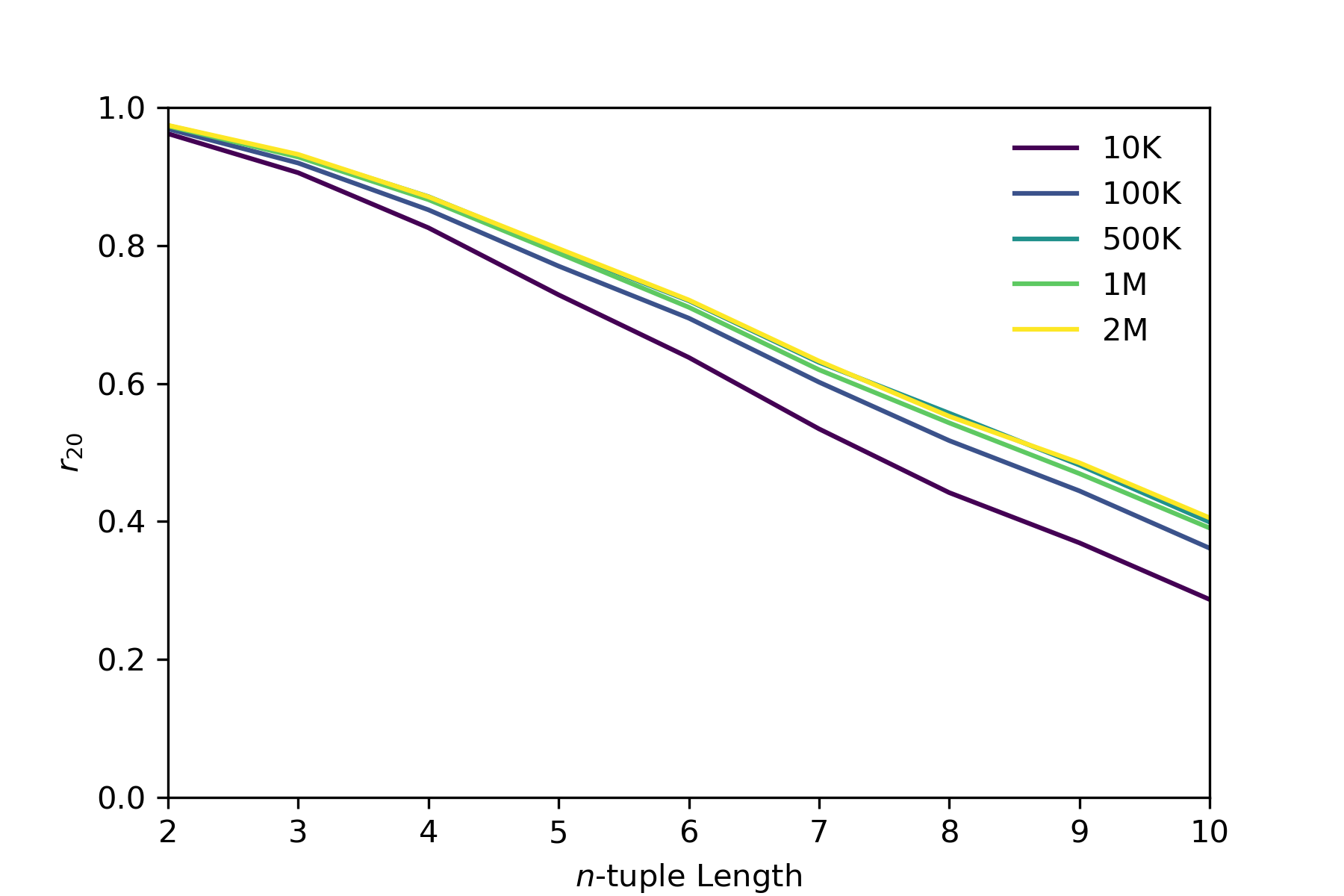}
\caption{sVAE performance for $l=7$ for varying synthetic training dataset sizes. For each training dataset size, two inferences are run with different random seeds, shown in solid and dashed lines for each training size.}
\label{fig:saturation}
\end{figure*}

The goal of the synthetic test with 1M training sequences in the main text is to eliminate out-of-sample error (overfitting) by using an extremely large training dataset. How large must the training dataset be to mostly eliminate out-of-sample error for the sVAE? In Fig. \ref{fig:saturation} we show tests for the $l=7$ sVAE for increasing training dataset sizes, finding that after 500K sequences the improvement in performance becomes small. This justifies our choice of using 1M synthetic training sequences, as there is little additional improvement to be gained by fitting to 2M sequences at the cost of increasingly prohibitive fitting time.

\section{Using sVAE as the synthetic target distribution} 

In the main text, our synthetic GPSM tests are performed using a Potts model as the synthetic target distribution. This means that the target distribution is constructed without higher-order interaction terms, and a Potts model is by definition well specified to fit data generated from this target distribution. Here, we show GPSM performance when the synthetic target distribution instead corresponds to a sVAE, which potentially generates data which cannot be fit by a model with only pairwise interaction terms.

In this section we take the synthetic target distribution to be described by the sVAE fit in the main text from 10K natural sequences. As described in the main text, this model potentially generates patterns of higher-order mutational covariation which cannot be fit by the Mi3 model. We then follow the same procedure as for our synthetic 1M test of the main text, but using this target distribution. We generate 1M sequences from the target sVAE distribution which we use as training data for each GPSM, that is for Mi3, sVAE, and Indep. We generate evaluation MSAs from each inferred model and compare it to evaluation MSAs generated from the target distribution, using our test statistics.

\begin{figure*}
\centering
\includegraphics[width=\textwidth]{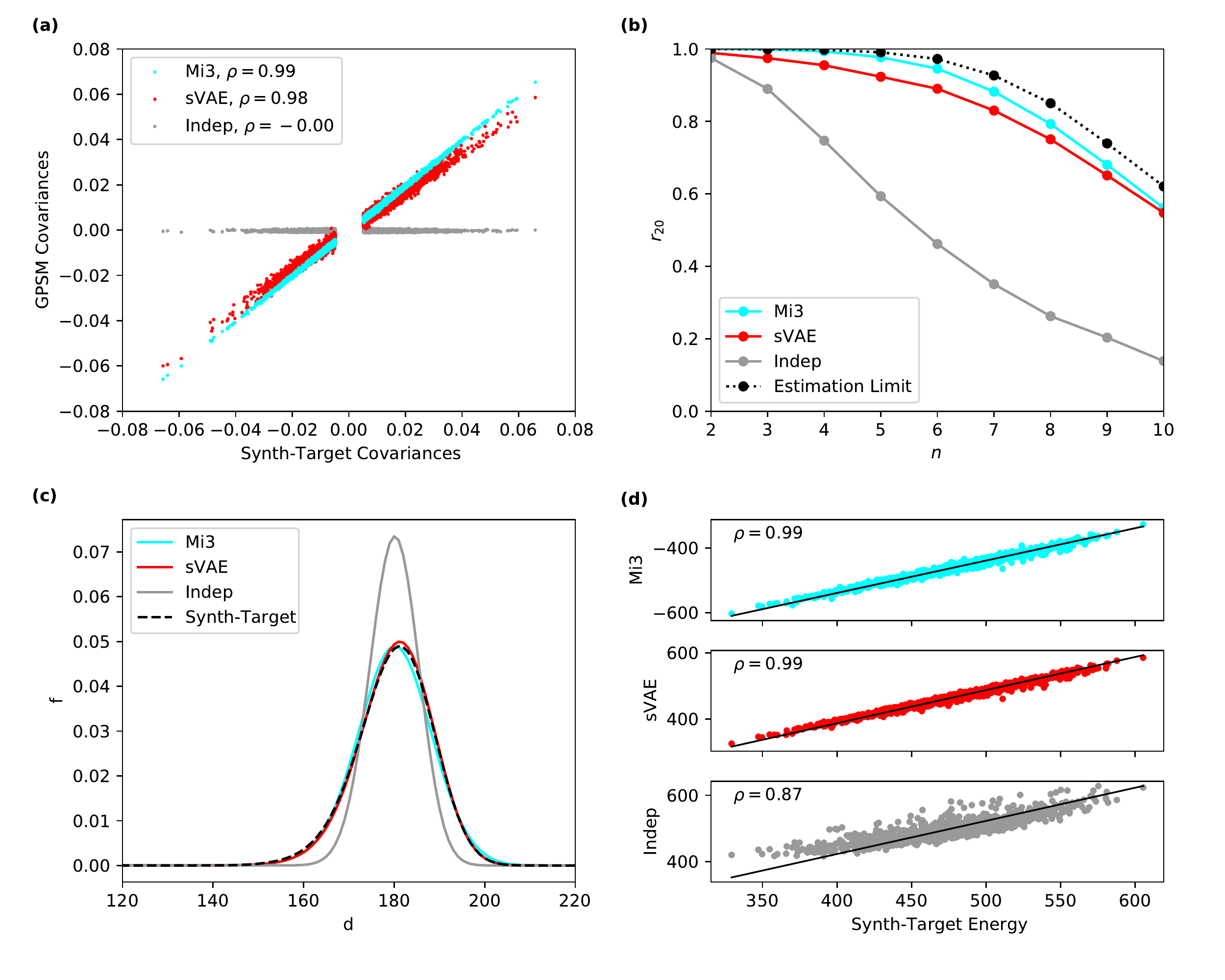}
\caption{Synthetic test of the performance of different GPSMs when the synthetic target distribution is specified by sVAE. \textbf{a} Pairwise covariance correlation scores, as in main text Figure 2a. \textbf{b} $r_{20}$ scores, as in main text Figure 2d. \textbf{c} Hamming distance distributions, as in main text Figure 3a. \textbf{d} Statistical energy scores, as in main text Figure 4 panels a, c, e.}
\label{fig:synthvae}
\end{figure*}

In Fig. \ref{fig:synthvae} we show MSA test statistics for the models fit to the sVAE target. We find that the performance of Mi3 fit to this target performs at least as well as the sVAE model fit to the same target. As in the main text 1M synthetic test, the correlation scores are estimated from 500K evaluation sequences from the target and each GPSM, the $r_{20}$ scores using 6M evaluation sequences, the Hamming distributions from 30K sequences, and the energies are evaluated for 1K sequences using 1000 Monte Carlo samples. For the $r_{20}$ test we measure the estimation limit due to the finite size of the evaluation MSAs by computing $r_{20}$ between two MSAs of size 6M generated from the target distribution. There is a small difference between the estimation error limit and the Mi3 result, which may be due to out-of-sample error due to the finite 1M training data, or due to specification error, and this difference is smaller than the difference of sVAE fit to the same target distribution (red). In sum, we interpret these tests to show that the Mi3 model is able accurately fit sVAE's target distribution.

\section{How higher-order covariation is represented by pairwise models}

One of the questions we address in the main text is whether different GPSMs are well specified to describe protein sequence variation, especially in the case of covariation of many positions in the sequence at once. Of particular interest is whether a model which explicitly includes only pairwise interactions, such as the Potts model, is sufficient to model higher order epistasis, or whether GPSMs with more complex functional forms, such as a VAE, are necessary.

For clarity, we give a brief example describing how Potts models can predict many patterns of higher-order covariation, meaning triplet and higher patterns of residue covariation, despite only modelling pairwise interactions. We illustrate this using a toy model describing sequences of length $L=3$  with two residue types A and B, with $2^3 = 8$ possible sequences, and show different forms of higher-order covariation which a pairwise model can and cannot fit. Detailed discussion and theoretical results suggesting why pairwise models are often sufficient to model many datasets have been provided by others\cite{Bialek2003, Bialek2006, Nemenman_2016}.

First, we show how such a Potts model generates triplet covariation. Consider a Potts model with parameters given by $J^{12}_{AA} = J^{23}_{AA} = -s$ for some interaction strength $s$ and all other field and coupling parameters are 0. This directly couples the character ``A'' between positions (1,2) and also positions (2,3). These interactions cause pairwise covariation between the directly coupled residues, and in the limit of large $s$ we find $C^{12}_{AA} = C^{23}_{AA} = 0.08$, or 8\%, but they also cause covariation between the indirectly coupled pair, as  $C^{13}_{AA} = 0.04$, or 4\%. Furthermore, this Potts model predicts three-body covariation, as can be seen by computing the three-body covariation terms found in cluster expansions in statistical physics given by 
\begin{equation}
    C^{123}_{\alpha\beta\gamma} = f^{123}_{\alpha\beta\gamma} 
    - f^1_\alpha C^{23}_{\beta\gamma} 
    - f^2_\beta C^{13}_{\alpha\gamma} 
    - f^3_\gamma C^{12}_{\alpha\beta}
    - f^1_\alpha f^2_\beta f^3_\gamma
\end{equation}
and we find that $C^{123}_{AAA} = 0.024$, or 2.4\%, which is nonzero. This shows that a Potts model generates and can fit higher-order covariation between sets of residues even though the interactions are only pairwise, as a result of indirect covariation through chains and loops of pairwise interactions. 

\begin{table}[ht]
\centering
\begin{tabular}[b]{c}
 AAA  \\
 ABB  \\ 
 BAB  \\
 BBA  \\ 
\end{tabular}
\caption{Example MSA following the XOR pattern.}
\label{tbl:xor_msa}
\end{table}

An example of MSA triplet statistics which a Potts model is mis-specified to describe is the XOR pattern in which the dataset is composed in equal proportions of copies of the four sequences shown in Table \ref{tbl:xor_msa}. These sequences follow the XOR function in boolean logic, so that the 3rd position is the XOR function applied to the first two positions. One can see that both the A and B residues have a 50\% probability at each position, and that for each pair of positions the probability of each of the four combinations AA, AB, BA, BB is 1/4. This means that the pairwise covariances $C^{ij}_{\alpha\beta} = 0.25 - 0.5 \times 0.5$ are all 0. Because there are no pairwise covariances, fitting a Potts model to this data will yield a model with no coupling terms, equivalent to an Indep model. Sequences generated from this (or any) Indep model have all three-body covariation terms equal to 0. However, the three-body covariations of the dataset are non-zero and $C^{123}_{AAA} = 0.125$. This shows how a Potts model fit to XOR data will fail to reproduce the correct three-body covariations. More generally, it will fail to model data which follows a boolean parity function, which generalizes the XOR function to longer strings, and is defined so that the last character is set to ``B'' if there are an odd number of ``B'' characters in the preceding sequence.

A motivation for the VAE is that it may potentially be able to model patterns of covariation such as the XOR pattern which a Potts model cannot. Whether a VAE is able to outperform the Potts model when fit to protein sequence data will depend on the prevalence of patterns such as XOR in the data which cannot be fit by a Potts model. If they are undetectable, the Potts model will be well specified and third order parameters are unnecessary. Our results with the natural dataset in the main text suggest no evidence that the Potts model is mis-specified to our dataset, as it is able to reproduce all the MSA statistics we tested up to the limits imposed by estimation and out-of-sample error.

\begin{figure*}
\centering
\includegraphics[width=\linewidth]{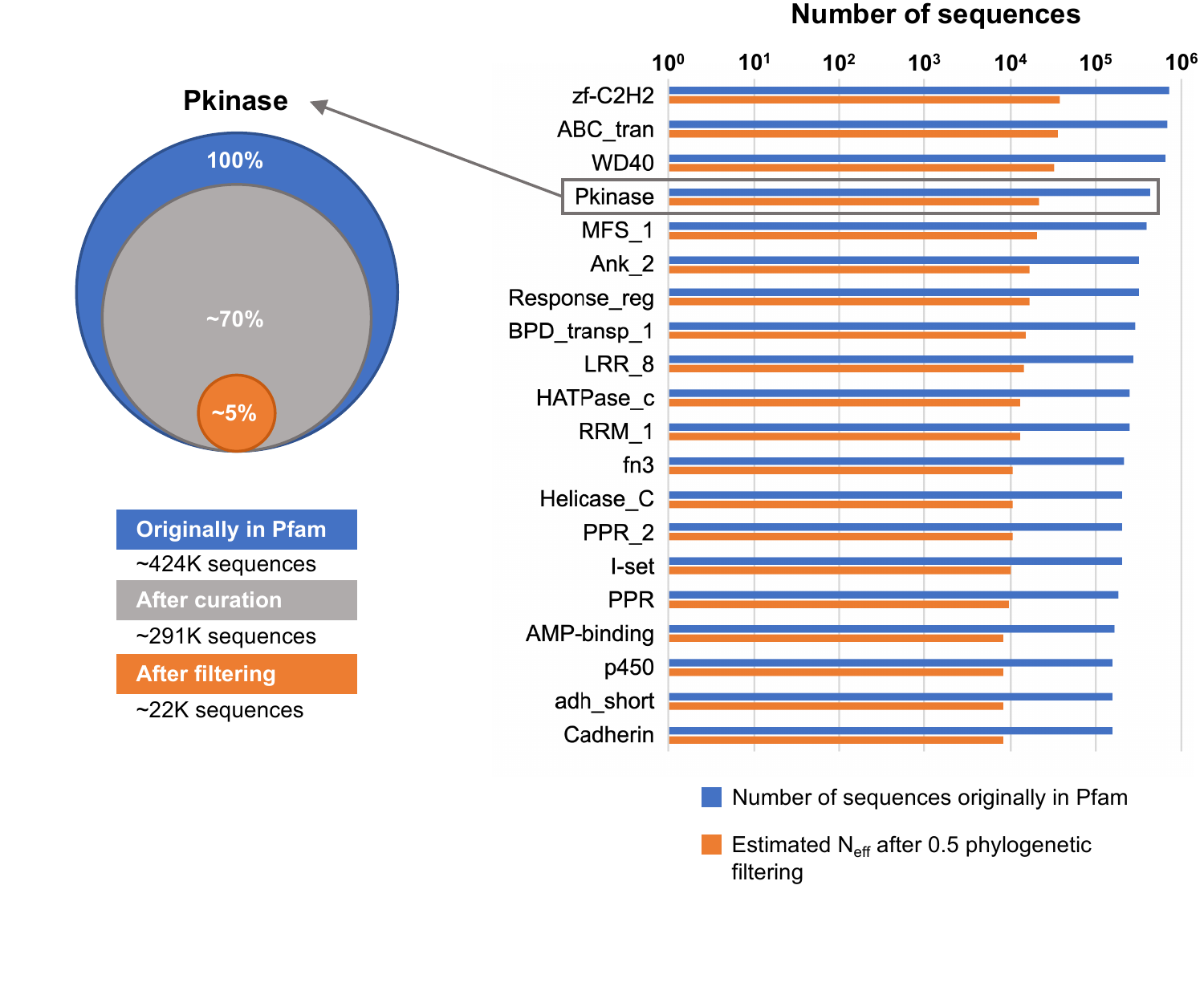}
\caption{\textbf{Pfam Top 20.} GPSMs trained on publicly available natural sequence data could be inherently data-starved. 
\textbf{Right} Log-scaled histogram of Pfam sequence frequencies. Sorted by the log-scaled number of sequences originally in Pfam (blue), the histogram shows estimated number of effective sequences $N_{\text{eff}}$ after phylogenetic filtering at the 0.5 similarity cutoff (orange). All estimates are based on the actual $N_{\text{eff}}$ for Pkinase, the fourth most frequent protein family and the one used in this work, which is $\sim$22K sequences, or $\sim$5\% of the total $\sim$424K Pkinase sequences in Pfam (left). Cadherin, the last entry (bottom), has $N_{\text{eff}} < 10^4$ (10K sequences), meaning that this must be the approximate upper-bound of $N_{\text{eff}}$ for GPSMs training on natural data outside the Pfam Top 20. Since all proteins outside the Pfam Top 20 must $N_{\text{eff}} < 10^4$, we chose 10K sequences as the lower limit of total training sequences for our synthetic analysis. \textbf{Left} Curation and phylogenetic filtering breakdown for Pfam Pkinase dataset. Of $\sim$424K Pkinase sequences in Pfam (blue), only $\sim$291K ($\sim$70\%) remained after curation (grey). This curated set was phylogenetically filtered at 0.5 similarity, resulting in $N_{\text{eff}}\sim$22K (orange), or 5\% of the original $\sim$424K.}
\label{fig:pfam}
\end{figure*}

\section{Analysis of estimation error}

When computing the $r_{20}$ scores we are able to quantify estimation error, as can be seen by the $r_{20}$ upper limit illustrated in Fig. \ref{fig:synthvae}b (black dotted line). Here we provide quantitative intuition for the behavior of the $r_{20}$ score as a function of the evaluation MSA size $N$, which explains the difficulty in eliminating estimation error entirely.

Consider a particular set of positions for which we estimate the frequency $f$ of each subsequence at those positions in the target distribution, based on a finite MSA of size $N$ generated from the target distribution, giving estimated marginals $\hat{f}$. We retain only the top twenty observed subsequences for use in the $r_{20}$ computation. The statistical variance in $\hat{f}$ caused by finite-sampling error will be  $f (1-f)/N$, following a multinomial sampling process, and we will approximate that all top 20 marginals have similar magnitude and we approximate this error as $\langle f \rangle(1-\langle f \rangle)/N$ for all twenty values, where $\langle f \rangle$ is the mean value of the top 20 marginals.

We can then approximate that the expected Pearson correlation $\rho^2$ between values estimated from two such MSAs will be $\rho^2 \approx \chi^2/(\chi^2 + \sigma^2)$ where $\chi^2$ is the variance in the values of the top 20 marginals (reflecting the variance of the ``signal''), and $\sigma^2 \approx \langle f \rangle (1-\langle f \rangle)/N$ is the statistical error in each value (representing the variance of the ``noise'').

$\langle f \rangle$ and $\chi$ are properties of the protein family being modelled, at each position-set, and do not depend on $N$. This invariant allows us to extrapolate, since if we solve for $\langle f \rangle (1-\langle f \rangle)/\chi^2 = N(1/\rho^2 - 1)$, the rhs should be invariant when we change the size of the dataset MSA from $N$ to $N_0$ or vice versa. If we estimate the rhs for a particular $N_0$ and measured $\rho_0$ numerically, we can solve for $\rho$ at higher $N$ since $N(1/\rho^2 - 1) = N_0(1/\rho^2_0 - 1)$, or
\begin{equation}
        N = N_0 \frac{\rho^2/(1-\rho^2)}{\rho_0^2/(1-\rho_0^2)}.
\end{equation}

The approximations we used to derive this formula will become more accurate for larger $N_0$. We have tested this formula by predicting the expected $r_{20}$ for MSAs of size $N$ by extrapolating based on the measured $r_{20}$ for MSAs of smaller size $N_0$, and find it is quite accurate.

This equation shows how extremely large MSAs can be required to reduce estimation errors when evaluating $r_{20}$, as the extrapolated $N$ diverges as $\propto 1/x$ as $x = 1 - \rho^2$ approaches 0. For instance, if with an MSA of 6 million sequences we obtain $r_{20} = 0.8$, then we would require 28.5 million sequence to obtain $r_{20} = 0.95$ and 148 million to reach $r_{20} = 0.99$.

\section{Typical natural sequence dataset MSA size}

The 10K sequence training datasets we use in the main text are meant to illustrate performance for typical protein family dataset sizes. The size of 10K sequences is the number of estimated effective sequences $N_{\text{eff}}$ remaining after curation and phylogenetic filtering for the 20th most frequent protein (Cadherin) in Pfam (Fig.\ref{fig:pfam}, right)\cite{el-gebali_pfam_2019}. Some of our measurements show significant out-of-sample error for Mi3 and the VAEs based on training sample size alone, suggesting that the vast majority of GPSMs training on natural data could be subject to the level out-of-sample error reported in our results.

In Pfams's Top 20 most frequent protein domains, ranked by total number of sequences, there are between $10^5$ and $10^6$ total sequences each (Fig.\ref{fig:pfam}, right). In this work, we use the 4th most frequent protein out of this ranking, Pkinase. After curation and phylogenetic filtering of the kinase, we retained only $N_{\text{eff}}\sim$22K, or $\sim$5\% of the original $\sim$424K kinase sequences (Fig.\ref{fig:pfam}, left). Extending this fraction of $\sim$5\% to the other Top 20 proteins, we estimate that $N_{\text{eff}}$ is capped at $\sim10^5$ (100K) for GPSMs trained on single domains, and that proteins outside the Top 20 can generally expect $N_{\text{eff}} < 10^4$ (10K). This tabulation of Pfam data demonstrates that, for the vast majority of proteins with publicly available natural sequence data, contemporaneous GPSMs must have approximately $N_{\text{eff}} < 10$K for training, validation, and testing.

\bibliography{supplement}